\pdfoutput=1
\documentclass[nohyperref,usenames,dvipsnames]{article}

\usepackage{microtype}
\usepackage{graphicx}
\usepackage{subfigure}
\usepackage{booktabs} % for professional tables

\usepackage{hyperref}
\usepackage[nointegrals]{wasysym}
\usepackage{xcolor}

\newcommand{\infor}{Informer}
\newcommand{\inten}{Intention}

\newcommand{\kvqs}{KVQ Models}

\newcommand{\fmtmatrix}[1]{\mathbf{#1}}
\newcommand{\fmtvector}[1]{\mathbf{#1}}
\newcommand{\fmtspace}[1]{\mathcal{#1}}
\newcommand{\fmtfunction}[1]{{#1}}
\newcommand{\fmtscalar}[1]{#1}
\usepackage[accepted]{icml2022}

\usepackage{amssymb}
\usepackage{mathtools}
\usepackage{caption}

\usepackage[capitalize,noabbrev]{cleveref}

\usepackage[bibliography=common]{apxproof}
\newtheoremrep{theorem}{Theorem}[section]
\newtheoremrep{proposition}[theorem]{Proposition}
\newtheoremrep{lemma}[theorem]{Lemma}
\newtheoremrep{corollary}[theorem]{Corollary}
\newtheorem{definition}[theorem]{Definition}

\newtheoremrep{remark}[theorem]{Remark}
\usepackage[textsize=tiny]{todonotes}

\icmltitlerunning{Exploring the Space of Key-Value-Query Models with Intention}

\begin{document}

\twocolumn[
\icmltitle{Exploring the Space of Key-Value-Query Models with Intention}



\icmlsetsymbol{equal}{*}

\begin{icmlauthorlist}
\icmlauthor{Marta Garnelo}{dm}
\icmlauthor{Wojciech Marian Czarnecki}{vl}

\end{icmlauthorlist}

\icmlaffiliation{dm}{DeepMind}
\icmlaffiliation{vl}{Voylab}

\icmlcorrespondingauthor{Marta Garnelo}{garnelo@google.com}

\icmlkeywords{Machine Learning, ICML}

\vskip 0.3in
]

% \appendixprelim{}


\printAffiliationsAndNotice{\icmlEqualContribution} % otherwise use the standard text.

% \begin{quote}
%     \textit{``You two suck. I leave you alone for two years and you degenerate to this!"}
%     \vspace{-15pt}
%     \begin{flushright}D. Balduzzi\end{flushright}
% \end{quote}

\begin{abstract}
Attention-based models have been a key element of many recent breakthroughs in deep learning.
Two key components of Attention are the structure of its input (which consists of keys, values and queries) and the computations by which these three are combined.
In this paper we explore the space of models that share said input structure but are not restricted to the computations of Attention.
We refer to this space as Keys-Values-Queries (KVQ) Space.
Our goal is to determine whether there are any other stackable models in KVQ Space that Attention cannot efficiently approximate, which we can implement with our current deep learning toolbox and that solve problems that are interesting to the community. 
Maybe surprisingly, the solution to the standard least squares problem satisfies these properties.
A neural network module that is able to compute this solution not only enriches the set of computations that a neural network can represent but is also provably a strict generalisation of Linear Attention.
Even more surprisingly the computational complexity of this module is exactly the same as that of Attention, making it a suitable drop in replacement.
With this novel connection between classical machine learning (least squares) and modern deep learning (Attention) established we justify a variation of our model which generalises regular Attention in the same way. 
Both new modules are put to the test an a wide spectrum of tasks ranging from few-shot learning to policy distillation that confirm their real-worlds applicability.
\end{abstract}

\section{Introduction}
\label{introduction}
In recent years we have witnessed a handful of extraordinary real world applications of deep learning. From models that can generate new text-conditioned images at the level of a professional illustrator~\citep{ramesh2022hierarchical} and the prediction of protein structures beyond what current experimental methods can achieve~\citep{jumper2021highly, verkuil2022language} to human-level chat bots that are able to generalise to new reasoning tasks ~\citep{brown2020language, schulman2022chatgpt}.
Being able to train unprecedentedly large models on increasingly larger data sets is a big part of these success stories. 
Another common ingredient, however, is the use of Attention modules (usually in the form of Transformers~\citep{vaswani2017attention}) as part of their architecture.

At a higher level, the input to an Attention block is divided into keys ($\fmtmatrix{K}$), values ($\fmtmatrix{V}$) and queries ($\fmtmatrix{Q}$) that are combined via multiplications to produce predictions. We will therefore refer to this family of models as \kvqs{} going forward. It is this multiplicative interaction between inputs - not found in the other aforementioned modules - that is hypothesised to be responsible for the increased expressive power of Attention models which could potentially explain their strong performance \citep{jayakumar2020multiplicative}. 

With this in mind, the goal of this paper is to address the following four-part question:

\fbox{\begin{minipage}{22.5em}
Are there other \kvqs{} that:
\begin{enumerate}
    \itemsep0em
    \item can not be efficiently approximated by Attention,
    \item can be stacked as a compositional computational block,
    \item can be easily implemented using existing deep learning tools and libraries, with the same same computational complexity,
    \item are useful for tasks of significance in the real world?
\end{enumerate}
\end{minipage}}

In order to study this question we search for \kvqs{} that can be neither represented nor approximated efficiently by Attention modules.  
In doing so we consider the solution to regularised least squares, one of the most widely used methods in data science~\cite{deaton1992understanding,krugman2009international}.
The size required by models like feed forward networks or even Attention to just approximate such a computation grows exponentially with the size of the problem.
We show that modules that include the solution to regularised least squares as part of their computation are indeed part of the KVQ Space and proceed to explore them as potential useful computational blocks for deep learning. 
As a nod to Attention and because least squares fitting has historically often been used to predict the future in time series~\cite{bianchi1999comparison} we call this module \emph{\inten{}}. 

Despite potentially appearing unrelated to Attention, we show that \inten{} is in actuality a generalisation of Linear Attention. 
With a minimal modification we obtain $\sigma$\inten{} that generalises regular Attention up to a rescaling factor. 
This connection between regularised least squares and Attention is something that to the best of our knowledge has not been studied before and opens up the door to new connections between classical machine learning tools and current deep learning approaches.

We demonstrate how \inten{} and $\sigma$\inten{} can be straightforwardly implemented as part of the deep learning toolbox with the same computational complexity as Attention (see Section~\ref{practical} for details). Furthermore we show how \inten{} can also represent other canonical machine learning problems such as linear discriminant analysis (LDA) and least-squares support vector machines (LS-SVMs). 

Finally, we provide experimental evidence to support our claims that \inten{} can be useful for a wide spectrum of relevant tasks such as: few-shot learning tasks (both regression as well as classification), policy distillation, point cloud distortion and outlier detection.

Our contributions can thus be summarised as:
\begin{enumerate}
    \item We introduce the notion of \kvqs{} and study alternative hypothesis spaces in this family of models.
    \item We identify least squares minimisers as an example of a computation that can't be approximated by Attention and is easy to implement using deep learning tools.
    \item We show that \inten{} is not only very closely related to Attention but can be seen as a generalisation of Linear Attention. 
    \item We are able to unify various seemingly unrelated meta-learning approaches using \inten{} as a framework which we outline in the related work section.
    \item We show that similarly to Attention, \inten{} can be stacked to obtain more powerful models which we term Informers.
\end{enumerate}

The proofs for all propositions and theorems can be found in the Appendix.

\section{Exploring the KVQ Space}
\label{exploring}

    %%%%%%% SOME FIRST PIECES %%%%%%%%%%%%\
    Let us start by formally defining the KVQ Space - a space of functions that learn to extract information from a collection of key-value pairs and apply it to a set of query points.

    \begin{definition}[KVQ Space]
    We define the $\mathcal{H}_\mathrm{KVQ}$ hypothesis space as a space of all functions 
    $$f:\mathbb{R}^{N \times d} \times \mathbb{R}^{N \times k} \times \mathbb{R}^{M \times d} \rightarrow \mathbb{R}^{M\times k}$$
    such that for every permutations $\rho_1, \rho_2$ 
    we have
    $$
    f(\rho_1 \circ \fmtmatrix{K}, \rho_1 \circ \fmtmatrix{V}, \rho_2 \circ  \fmtmatrix{Q} ) = \rho_2 \circ f(\fmtmatrix{K}, \fmtmatrix{V}, \fmtmatrix{Q}).
    $$
    \end{definition}

    There are many specific incarnations of what is called Attention~\cite{att0,att1,att2,att3}. For simplicity we focus on the form of Attention that is used by Transformers~\citep{vaswani2017attention} but analogous results can be shown for many other variants.

    \begin{definition}[Attention Space]
    We define $\fmtspace{H}_\mathrm{att}$ as the space of all functions of form
    $
    \sigma( \fmtmatrix{Q} \fmtmatrix{\theta}_1 \fmtmatrix{K}') \fmtmatrix{V} \fmtmatrix{\theta}_2
    $
    where $\sigma$ is a row-wise softmax and $\theta$ are parameter matrices of appropriate sizes.
    \end{definition}
    
    And analogously we can define a space spanned by Transformers. For simplicity we omit LayerNorm layers as those do not add any expressivity but make the math more complex.

    \begin{definition}[Transformer Space]
    We define $\fmtspace{H}_\mathrm{trans}$ as a space of all functions (potentially iteratively applied) of form
    $
    \fmtfunction{e}^{(i)}_{o}(\fmtmatrix{X} + f^{(i)}_\textrm{att}(\fmtmatrix{X},\fmtmatrix{X},\fmtmatrix{X}))
    $
    where each $f^{(i)}_\textrm{att} \in \fmtspace{H}_\textrm{att}$, $\fmtmatrix{X} = [\fmtmatrix{K}; \fmtmatrix{V}; \fmtmatrix{Q}]$ enriched with positional encoding,  $\fmtfunction{e}^{(i)}_{o}$ is an embedding MLP, and the final layer extracts only $M$ predictions. 
    \end{definition}
    
    \begin{figure}[ht]
    \begin{center}
    \centerline{\includegraphics[width=0.7\columnwidth]{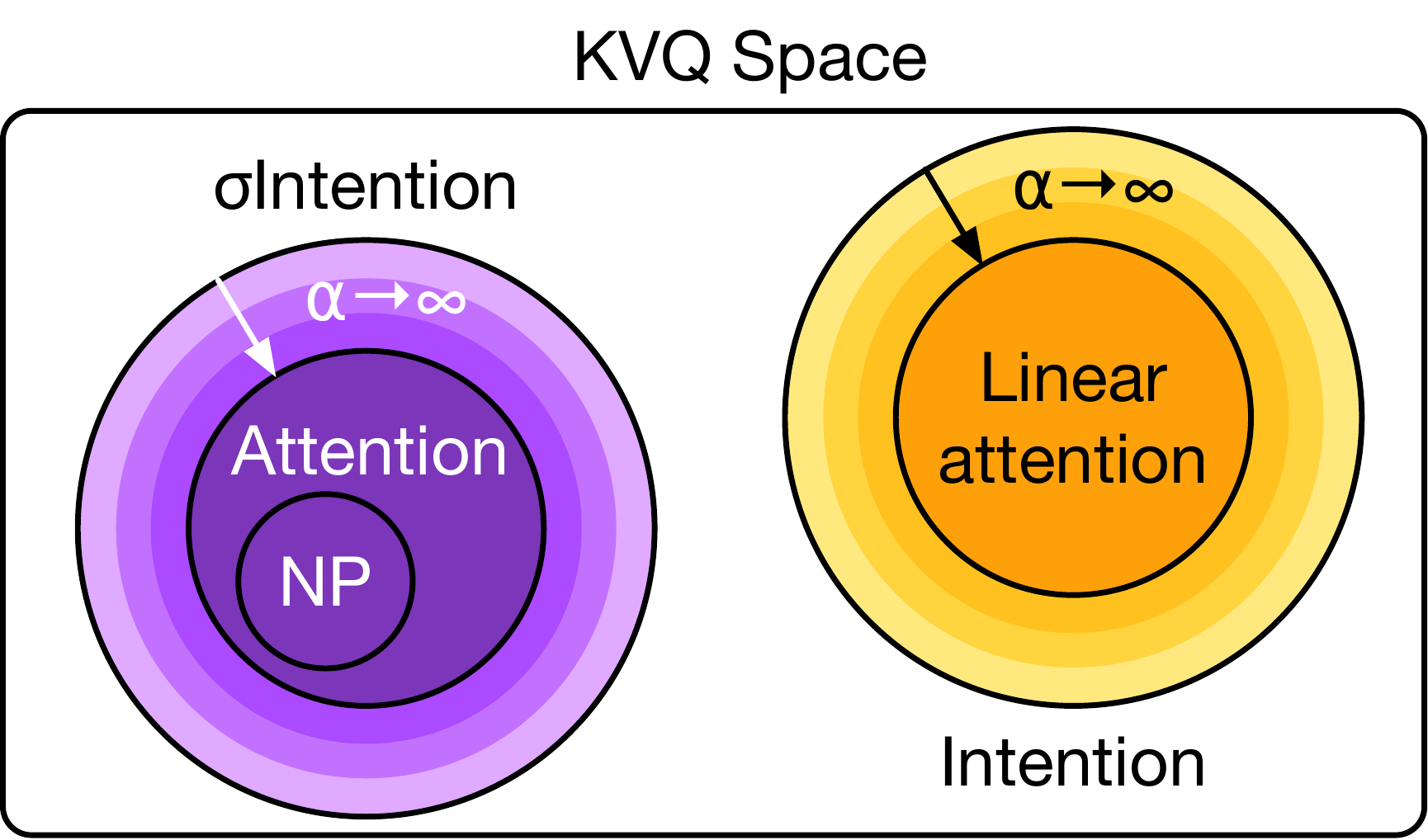}}
    \caption{Diagram representing how Intention and $\sigma$Intention relate to Attention and Linear Attention in the KVQ Space.}
    \label{fig:spaces}
    \end{center}
    \vskip -0.2in
    \end{figure}
    It is easy to see that $\fmtspace{H}_\mathrm{att} \subset \fmtspace{H}_\mathrm{KVQ}$ and $\fmtspace{H}_\mathrm{trans} \subset \fmtspace{H}_\mathrm{KVQ}$ (depicted in Figure~\ref{fig:spaces}). 
    Furthermore this part of the KVQ Space has numerous important properties:
    it is has been shown empirically to provide state of the art results in a plethora of applications, ranging from AI for video games~\cite{berner2019dota}, through image and audio synthesis~\cite{saharia2022photorealistic,wang2023neural} to structural biology~\cite{jumper2021highly}. 
    It can be seen as a quite simple, generic way of processing sets~\cite{lee2019set}, sequences~\cite{vaswani2017attention} and even graphs~\cite{kim2022pure}. 
    Due to its incorporation of multiplicative interaction it also enables efficient expression of important building blocks of algorithms, such as conditional execution or similarity between pairs of points, both of which require exponentially many parameters when using regular networks~\cite{jayakumar2020multiplicative}. 
    Finally, they can express heavily localised computations, which can lead to very non smooth behaviour (in Lipschitz terms)~\cite{kim2021lipschitz}.
    
    With all these properties it is reasonable to ask whether there are any parts of the KVQ Space that Attention does not cover. Are there functions in this space that 1. Attention can neither represent nor efficiently approximate and 2. at the same time are relevant for real life applications?
    
    To motivate both parts of this question let us look at two counter examples that only satisfy one of the two requirements each. 
    
    Example 1: consider $f_\mathrm{erf}(\fmtmatrix{K},\fmtmatrix{V},\fmtmatrix{Q}):= \mathrm{erf}(\fmtmatrix{Q}\fmtmatrix{K}')\fmtmatrix{V}$ where $\mathrm{erf}(z) = \frac{2}{\sqrt{\pi}}\int_{0}^{z}\exp^{-t^2}dt$ is applied pointwise. 
    % It is easy to show that indeed an Attention-based model can't represent such function
    \begin{propositionrep}
    $f_\mathrm{erf} \in \fmtspace{H}_\mathrm{KVQ} \setminus \fmtspace{H}_\mathrm{att}$.
    \end{propositionrep}
    \begin{appendixproof}
    Let's consider $\fmtmatrix{K} = [[\kappa]]$, $\fmtmatrix{V} = [[1]]$, $\fmtmatrix{Q} = [[1]]$ 
    therefore
    $f_\mathrm{erf}(\fmtmatrix{K},\fmtmatrix{V},\fmtmatrix{Q}) = f_\mathrm{erf}(\kappa)$.
    According to Liouville’s theory of Differential Algebra theorem~\cite{liouville1833premier} $f_\mathrm{erf}(\kappa)$ cannot be represented with addition, multiplication, exponents, or logarithms.
    Both Attention and Transformers on the other hand are composed exclusively of these computations and therefore cannot represent $f_\mathrm{erf}$.
    \end{appendixproof}
    % is therefore an example of a function that can not be represented by Attention, so it
    As such, $f_\mathrm{erf}$ satisfies the first condition. However, the function itself is quite exotic and the need for such a computation in a neural network is hard to justify given that it is very similar to a sigmoid function~\citep{vinyals2019grandmaster}, so it does not satisfy the second. As a result, while it does expand the space of functions representable by current Deep Learning approaches it's hard to justify its use case.
    
    Example 2: Inversely, there are KVQ Models that are more common and useful but that Attention \emph{is} able to represent.
    The computation of Neural Processes~\citep{garnelo2018conditional} is one such example.
    \begin{propositionrep}
    A Transformer can exactly represent NP computation, but simple Attention module cannot.
    \end{propositionrep}
    \begin{appendixproof}
    Neural process (NP) can be seen as a collection of functions of form
    $$
    \fmtfunction{f}_\textrm{NP}(\fmtmatrix{K},\fmtmatrix{V},\fmtmatrix{Q}):= \fmtfunction{e}_o\left (\fmtmatrix{Q}, \tfrac{1}{N} \sum_i \fmtfunction{e}_c([ \fmtmatrix{K}_i;\fmtmatrix{V}_i ] ) \right ),
    $$
    where $\fmtfunction{e}_\cdot$ are MLP-based embeddings. Since Attention is a linear operator with respect to $\fmtmatrix{V}$ and NP is not, it clearly cannot represent it.
    
    With a Transformer, let us put $\fmtmatrix{K} = \boldsymbol{1}$ (a matrix of 1s) and as a result Attention computes an average of values. Using learnable embeddings for keys, values and queries we can get
    $$
    \fmtfunction{f}_\mathrm{att}(\boldsymbol{1},[\fmtmatrix{K}; \fmtmatrix{V}],\fmtmatrix{Q}) = \tfrac{1}{N} \sum_i \fmtfunction{e}_c([ \fmtmatrix{K}_i;\fmtmatrix{V}_i ] ).
    $$
    With a Transformer skip connection we get
    $$
    \fmtfunction{e}_o\left ( [\fmtmatrix{K}; \fmtmatrix{V}; \fmtmatrix{Q}] + \tfrac{1}{N} \sum_i \fmtfunction{e}_c([ \fmtmatrix{K}_i;\fmtmatrix{V}_i ] ) \right ).
    $$
    The only difference is thus concatenation over addition. However this can be simply represented by making sure that $\fmtfunction{e}_o$ has 2 times bigger first layer, and $\fmtmatrix{Q}$ occupies first half, while $\fmtfunction{e}_c$ the latter, thus simulating concatenation.
    \end{appendixproof} 
    
    As such, a Neural Process, while useful in its computations \emph{can} be represented by a Transformer, so it does not expand the current Deep Learning toolbox.
    
    A computation that an Attention-based model cannot represent \emph{and} that is useful (having become one of the most commonly used techniques in data science) is the solution to a standard regularised least squares minimisation problem.
    $$
    f_\mathrm{ls}(\fmtmatrix{K},\fmtmatrix{V},\textbf{Q}) := 
    \fmtmatrix{Q} [\fmtmatrix{K} \fmtmatrix{K}' + \alpha \fmtmatrix{I}]^{-1} \fmtmatrix{K}' \fmtmatrix{V}.
    $$ 
    
    \begin{theoremrep}
    Neither an Attention-based model nor a Transformer-based one can represent least squares fit.
    \end{theoremrep}
    \begin{appendixproof}
    Let us consider $N=M=d=k=1$ and $\fmtmatrix{K}=[[\kappa]], \fmtmatrix{V}=[[1]], \fmtmatrix{Q}=[[1]]$. We then have $\fmtfunction{f}_\mathrm{ls}(\fmtmatrix{K}, \fmtmatrix{V}, \fmtmatrix{Q}) =  \tfrac{1}{\kappa}$ and consequently
    $\lim_{\kappa \rightarrow 0} \fmtfunction{f}_\mathrm{ls}(\fmtmatrix{K}, \fmtmatrix{V}, \fmtmatrix{Q}) = \infty$
     and in particular for $\kappa = 0$ computation is ill defined. However, both Attention's and Transformer's computation is well defined for any parameters values, with inputs like above.
    \end{appendixproof}
    
\begin{figure}[ht]

\begin{center}
\centerline{\includegraphics[width=\columnwidth]{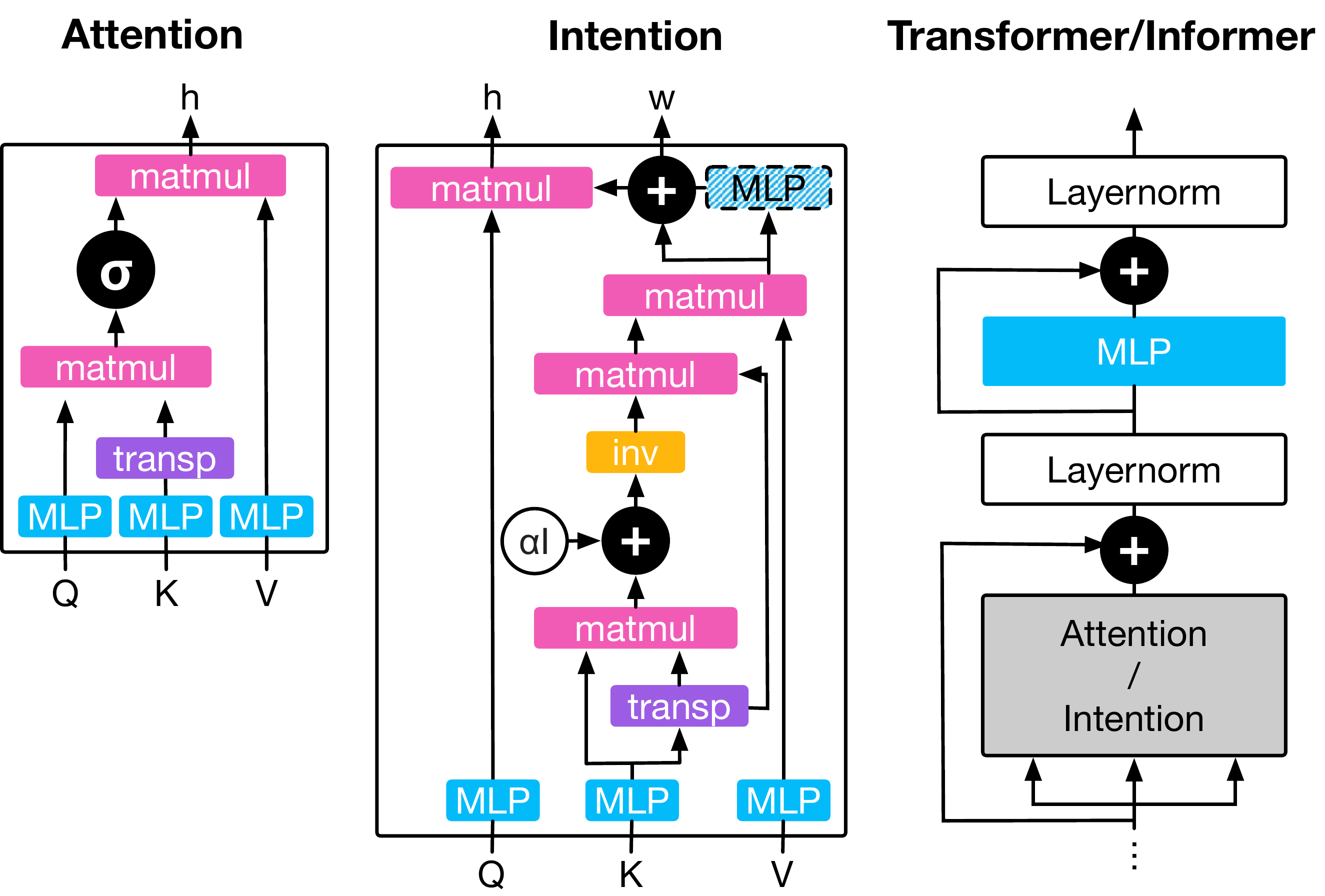}}
\caption{Diagrams of the models. From left to right: Attention head, Intention head and one of the layers of a Transformer/Informer. The latter depends on the module that is in the gray box. \emph{transp} indicates a matrix transpose operation and \emph{inv} inverse operation. A more extensive version can be found in the Appendix in Section~\ref{app:diagram}.}

\label{fig:atchs}
\end{center}
\vskip -0.2in
\end{figure}
    
    Similarly to the Attention Space we can now define an Intention Space that includes the models that able to represent least squares fit.
    
    \begin{definition}[Intention Space]
    We define $\fmtspace{H}_\mathrm{int}$ as a space of all functions of form
    $$
    \fmtmatrix{Q} \fmtmatrix{\theta}_1 [(\fmtmatrix{K}\fmtmatrix{\theta}_2)'(\fmtmatrix{K}\fmtmatrix{\theta}_2) + \alpha \fmtmatrix{I} ]^{-1} (\fmtmatrix{K}\fmtmatrix{\theta}_2)' \fmtmatrix{V} \fmtmatrix{\theta}_3
    $$
    where $\theta$ are parameter matrices of appropriate sizes, and $\alpha \geq 0$ is a covariance smoothing parameter.
    \end{definition}

\begin{figure}[ht]

\begin{center}
\centerline{\includegraphics[width=\columnwidth]{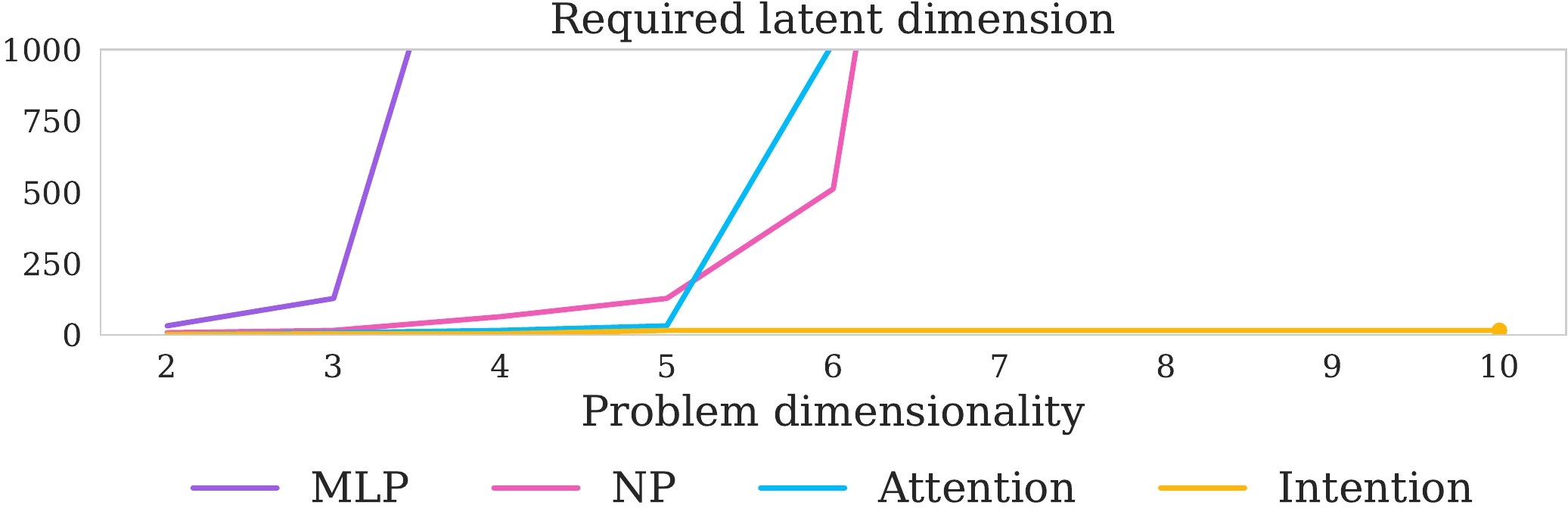}}
\caption{Comparison of the latent size dimension required by each of the models to efficiently approximate least squares regression computation in $\mathbb{R}^d$ space.}
\label{fig:scaling}
\end{center}
\vskip -0.2in
\end{figure}

Attention-based models, and especially Transformers, are universal approximators. Consequently it is natural to ask whether their inability to represent regularised least squares exactly is indeed an issue. We argue, that one needs to at least be able to \emph{efficiently} approximate a function for it to be a good surrogate solution. In order to test whether Attention is able to at least efficiently approximate regularised least squares we run the following experiment: we generate a data set of points obtained with different linear regression parameters. % with increasing data dimension. 
At every iteration we provide each model with 10 pairs of observations $\fmtmatrix{K}$, $\fmtmatrix{V}$ generated using one set of parameters and ask it to predict the parameters themselves.  
We test this for increasingly larger data dimensions  and determine what the smallest latent size of each architecture needs to be able to learn this data.
In Figure~\ref{fig:scaling} we can see that MLP, NP and Attention require exponentially many dimensions wrt. the input dimensionality, which confirms that they are not \emph{efficient} at approximating it.

It is natural to ask if there is some deeper connection between Attention and Intention modules. 
One such connection can be drawn from recent work of~\citet{von2022transformers} which shows, that internally Attention modules can be seen as a model of linear regression gradient updates. 
Each layer of Attention becomes then a single gradient update, and a stack of them can learn arbitrary dynamics. 
From this perspective one can think of an Intention layer as Attention behaviour in the limit, directly computing the minimum of the least squares problem, rather than doing so iteratively.

Furthermore, we can see that depending on the covariance smoothing strength, Intention interpolates its behaviour from minimum seeking to a simple Attention-like update.
\begin{theoremrep}
\label{linatt}
As the smoothing strength of the covariance estimator goes to infinity, an Intention module converges to a Linear Attention module up to a rescaling of queries.
\end{theoremrep}
\begin{appendixproof}
\begin{equation}
\begin{aligned}
\lim_{\alpha \rightarrow \infty}
f_{\textrm{int}}(\fmtmatrix{K},\fmtmatrix{V},\alpha \fmtmatrix{Q}) &= \lim_{\alpha \rightarrow \infty} \alpha \fmtmatrix{Q} [ \fmtmatrix{K}'\fmtmatrix{K} + \alpha \fmtmatrix{I} ]^{-1} \fmtmatrix{K}' \fmtmatrix{V}\\ 
&= \lim_{\alpha \rightarrow \infty} \fmtmatrix{Q} [ \tfrac{1}{\alpha} \fmtmatrix{K}'\fmtmatrix{K} + \fmtmatrix{I} ]^{-1} \fmtmatrix{K}' \fmtmatrix{V}\\ 
&=  [\fmtmatrix{Q} \fmtmatrix{K}'] \fmtmatrix{V} = f_{\textrm{lin-att}}(\fmtmatrix{K},\fmtmatrix{V},\fmtmatrix{Q})
\end{aligned}
\end{equation}
\end{appendixproof}
Directly from this relation, we can also see that if we were to define $\sigma$Intention as
$$
f_{\sigma \textrm{int}}(\fmtmatrix{K},\fmtmatrix{V},\fmtmatrix{Q}) = \sigma(\fmtmatrix{Q} [ \fmtmatrix{K}'\fmtmatrix{K} + \alpha \fmtmatrix{I} ]^{-1} \fmtmatrix{K}' ) \fmtmatrix{V}
$$
we would also get analogous theorem
\begin{theoremrep}
\label{sigmaint}
As the smoothing strength of the covariance estimator goes to infinity, a $\sigma$Intention module converges to an Attention module up to a rescaling of queries.
\end{theoremrep}
\begin{appendixproof}
\begin{equation}
\begin{aligned}
\lim_{\alpha \rightarrow \infty}
&f_{\sigma \textrm{int}}(\fmtmatrix{K},\fmtmatrix{V}, \alpha \fmtmatrix{Q})\\ &= \lim_{\alpha \rightarrow \infty} \sigma( \alpha \fmtmatrix{Q} [ \fmtmatrix{K}'\fmtmatrix{K} + \alpha \fmtmatrix{I} ]^{-1} \fmtmatrix{K}' ) \fmtmatrix{V}\\ 
&= \sigma( \fmtmatrix{Q} \fmtmatrix{K}' ) \fmtmatrix{V} = f_{\textrm{att}}(\fmtmatrix{K},\fmtmatrix{V},\fmtmatrix{Q})
\end{aligned}
\end{equation}
\end{appendixproof}
\begin{figure}
    \centering
    \includegraphics[width=\columnwidth]{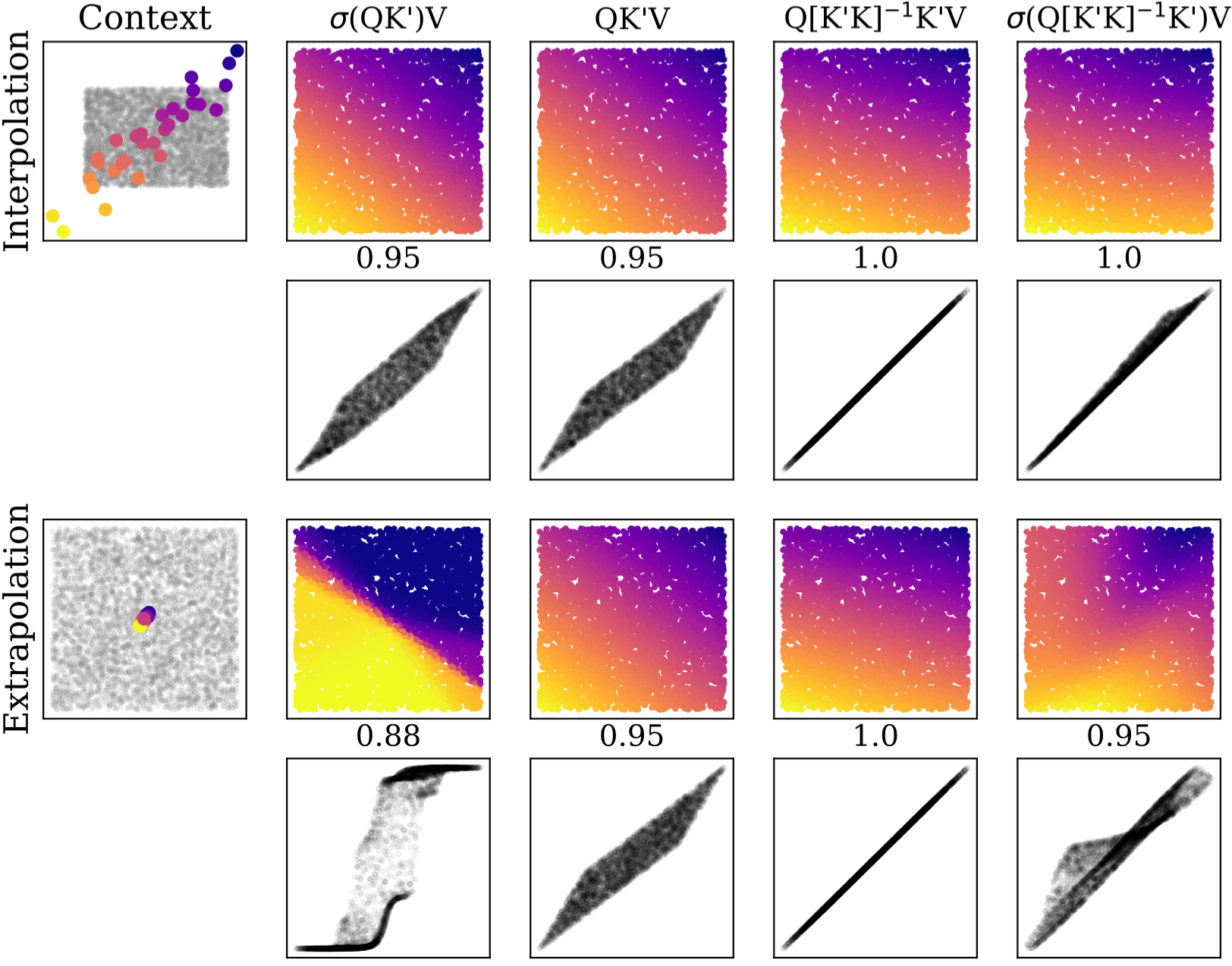}
    \caption{Comparison between the computations of (from left to right) Attention, Linear Attention, \inten{} and $\sigma$\inten{} when applied to the 2D linear regression problem. First and third rows show the results of interpolation and extrapolation respectively, based on context points present in the upper left corner. Second and fourth rows show the correlation between real value (x-axis) and predicted one (y-axis). Titles include Pearson correlation coefficient between prediction and ground truth.}
    \label{fig:attvsint}
\end{figure}
We depict the relation between these models in KVQ Space in Figure~\ref{fig:spaces} and a diagram of the computations of Attention and Intention in Figure~\ref{fig:atchs}. In Figure~\ref{fig:attvsint} we show how each of the models behave when carrying out a linear regression task. We sample context points from a 2D skewed Gaussian and attach target values through a random linear mapping. We then use the computations of Attention, Linear Attention, Intention and $\sigma$Intention to predict values over a square of the input space. We would like to point out that the predictions are simple carried out by applying the computations to the data directly, these are not trainable models. We can see that both Intention and $\sigma$Intention do perfectly well in the interpolation scenario, while both forms of Attention fail to capture the linear relation closely. The effect is arguably even stronger once we look at the extrapolation results where our query vectors are much larger than the keys (more details in Section~\ref{app:jaxu} of the Appendix).

Finally, one could ask if the solution to least squares regression can't be efficiently approximated by simply stacking Transformer layers. While we cannot provide a very strict limiting argument at the time, we analyse this idea as follows: the convergence rate of gradient descent is linear for quadratics~\cite{mitcourse}, and its exact speed depends on the \emph{conditioning number} (the ratio between the largest and the smallest eigenvalue of the covariance matrix). 
If we assume the solution comes from a standard Gaussian, this means that the initial optimality gap is $d$, and to reduce it by 0.1 we would need just one step (and thus one layer of Transformer) if the conditioning number was 1.1 (which would correspond to a trivial problem where the covariance is almost the identity). If the conditioning number reached $10$~\cite{mitcourse} and the latent space was of size $d=1000$ we would need roughly $6 \times 4=24$ stacked Transformer layers to reduce the error to merely $0.1$. Note that this is all assuming perfect selection of learning rate (exactly one over the Lipschitz constant) which in itself is a nontrivial mapping to represent.

With all of the results above we argue that there is a benefit to adding a least-squares solver neural network module to our deep learning toolbox. In the next section we describe the practical implementation of such a module, that is at least as expressive as $\fmtspace{H}_\textrm{int}$.

\section{Implementing \inten{} and \infor}
\label{implementing}

In this section we provide more details about a practical implementation of the Intention module within a deep learning framework.
We will use $\fmtfunction{e}_\cdot$ to denote a learnable MLP-based embedding.
% , and for simplicyt $\fmtmatrix{E}_\fmtmatrix{X} := \fmtfunction{e}_\fmtmatrix{X}(\fmtmatrix{X})$

First, we define an Intention network module as:
\begin{definition}[Intention module]
\begin{equation}
\begin{aligned}
\fmtfunction{h}_\mathrm{int}( \fmtmatrix{K}, \fmtmatrix{V}, \fmtmatrix{Q} ) &:= \fmtmatrix{E}_\fmtmatrix{Q} \fmtfunction{w}( \fmtmatrix{K}, \fmtmatrix{V} )\\
    \fmtmatrix{E}_\fmtmatrix{Q} &:= \fmtfunction{e}_\fmtmatrix{Q}(\fmtmatrix{Q}) \\
    \fmtfunction{w}( \fmtmatrix{K}, \fmtmatrix{V} ) & :=
    % \fmtfunction{e}_\fmtmatrix{O}(
    \fmtfunction{e}_\fmtvector{w}[ \fmtfunction{\Sigma}(\fmtmatrix{E}_\fmtmatrix{K}) ^{-1}\fmtmatrix{E}_\fmtmatrix{K}' \fmtmatrix{E}_\fmtmatrix{V}] \\
    \fmtmatrix{E}_\fmtmatrix{K} &:= \fmtfunction{e}_\fmtmatrix{K}(\fmtmatrix{K}) \\
    \fmtmatrix{E}_\fmtmatrix{V} &:= \fmtfunction{e}_\fmtmatrix{V}(\fmtmatrix{V}) \\
\end{aligned}
\end{equation}
where $\Sigma$ is a function providing some form of covariance estimation. By default we use $\Sigma(\fmtmatrix{X}) := \fmtmatrix{X}'\fmtmatrix{X} + \alpha \fmtmatrix{I}$ where $\alpha$ is either a hyperparameter or a learnable parameter.
\end{definition}
It is easy to see that as long as each $\fmtfunction{e}$ can represent the identity mapping the hypotheses space of this module contains $f_\mathrm{ls}$.% is at least as big as the $\fmtspace{H}_\mathrm{ls}$.

Similarly to Attention, Intention can also be applied in a Self-Intention mode, where we simply put
$
\fmtfunction{w}(\fmtmatrix{X}) := \fmtfunction{w}(\fmtmatrix{X},\fmtmatrix{X},\fmtmatrix{X}).
$

And the same way Self-Attention gets wrapped into a Transformer we can do the same with Self-Intention that becomes an Informer. This allows us to use Intention modules in a stackable way, if our problem requires this form of computation. In Section~\ref{informer} we show that indeed for some tasks the depth of Informers improves performance.
Similarly to Transformer we also can utilise multiple heads by splitting our embeddings space into subsets, performing computation and then merging it. 

In the above formulation Intention module can be seen as performing linear regression in latent space if values are continuous, or least squares support vector machines computation if values are categorical~\cite{suykens1999chaos}. 

\subsection{Practical considerations}
\label{practical}
The computational complexity of the inference is cubic in the number of rows of the $\Sigma$
matrix. In the above implementation this corresponds to the latent space dimensionality. 
If this becomes large we can use a dual formulation, where the complexity becomes cubic in the number of context points instead: 
$
\fmtmatrix{Q} (\fmtmatrix{K}'\fmtmatrix{K})^{-1}\fmtmatrix{K}'\fmtmatrix{V} =
[\fmtmatrix{Q} \fmtmatrix{K'}] (\fmtmatrix{K}\fmtmatrix{K}')^{-1} \fmtmatrix{V},
$
or one can use an additional mapping $\phi(\fmtmatrix{X}) := \fmtmatrix{X} \fmtmatrix{K}'(\fmtmatrix{K}\fmtmatrix{K}')^{-1/2} $ applied to both $\fmtmatrix{E}_\fmtmatrix{K},\fmtmatrix{E}_\fmtmatrix{Q}$. Both these approaches are equivalent, with the first being easier to implement, and second allowing us to also incorporate an arbitrary kernel function $\fmtspace{K}$ and obtain \begin{equation}
    \phi_\fmtspace{K}(\fmtmatrix{X}) := \fmtspace{K}(\fmtmatrix{X}, \fmtmatrix{K})\fmtspace{K}(\fmtmatrix{K}, \fmtmatrix{K})^{-1/2}.
    \label{eq:kerneltrick}
\end{equation} The latter can be further used to reduce computational complexity by using Nystroem method of kernel approximation~\cite{williams2000using,czarnecki2017extreme}.

%%%%%%%%%%%%%%%%%%%
While cubic complexity might seem large, it is important to note that this is the exact same cost as that of Attention. We often focus on Attention's quadratic memory complexity, but when it comes to compute its complexity is effectively cubic in either hidden dimension or in the number of samples.
\begin{theoremrep}
The asymptotic computational complexity of Intention is exactly the same as that of Attention and equals to $\mathcal{O}(N^2 d)$.
\end{theoremrep}
\begin{appendixproof}
For simplicity we analyse Self-Attention and Intention and we ignore the multiplication with values that is carried out in both as they share that computation and treat it the same way. Lets assume we have embedded queries into a $\fmtmatrix{Q} \in \mathbb{R}^{N \times d}$ matrix and keys into a $\fmtmatrix{K} \in \mathbb{R}^{N \times d}$.

\textbf{Complexity of Attention:} $f_\textrm{att}(\fmtmatrix{Q}, \fmtmatrix{K}) =( \fmtmatrix{Q} \fmtmatrix{K}' )$ is $\mathcal{O}(N^2d)$ (as this is the complexity of multiplying these two matrices).

\textbf{Complexity of Intention:} 

\begin{itemize}
    \item If $d < N$:
    \begin{itemize}
        \item $f_\textrm{int}(\fmtmatrix{Q}, \fmtmatrix{K}) =( \fmtmatrix{Q} (\fmtmatrix{K}'\fmtmatrix{K})^{-1}\fmtmatrix{K}') = \fmtmatrix{Q}\fmtmatrix{Z}^{-1}\fmtmatrix{K}'$
        \item Computing $\fmtmatrix{Z}^{-1}$ is $\mathcal{O}(N^2d)$: $\fmtmatrix{Z}$ is of shape $d\times~d$ so its inversion is $\mathcal{O}(d^3)$ which is $\mathcal{O}(N^2d)$ because $d < N$.
        \item All remaining operations are $\mathcal{O}(N^2d)$
        \item The whole computation is therefore $\mathcal{O}(N^2d)$, just like Attention.
        \end{itemize}
        
    \item If $d \geq N$:
    \begin{itemize}
        \item $f_\textrm{int}(\fmtmatrix{Q}, \fmtmatrix{K}) =( \fmtmatrix{Q} \fmtmatrix{K}'(\fmtmatrix{K}\fmtmatrix{K}')^{-1} = \fmtmatrix{Q}\fmtmatrix{K}'\fmtmatrix{Z}^{-1}$
        \item Computing $\fmtmatrix{Z}^{-1}$ is $\mathcal{O}(N^2d)$: $\fmtmatrix{Z}$ is of shape $N\times~N$ so its inversion is $\mathcal{O}(N^3)$ which is $\mathcal{O}(N^2d)$ because $d \geq N$.
        \item All remaining operations are $\mathcal{O}(N^2d)$
        \item The whole computation is therefore $\mathcal{O}(N^2d)$, just like Attention.
        \end{itemize}
\end{itemize} 

There are also asymptotically slightly faster ways of computing dot products, that decrease cubic time to roughly, however they are not really used in practise (as their constants are extremely bad), and even if they were, similar tricks can be applied to inversion too.
\end{appendixproof}

\begin{figure}[ht]
\begin{center}
\centerline{\includegraphics[width=\columnwidth]{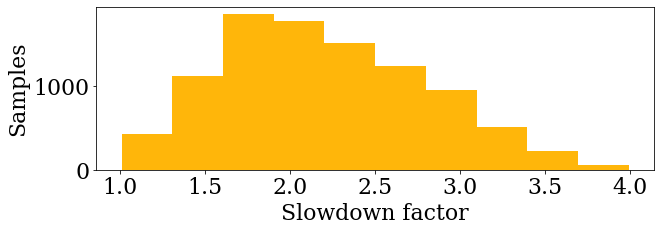}}
\caption{Slowdown factor of Intention over Attention. Distribution of the slowdown over 1809 independent repetitions.}
\label{fig:slowdown}
\end{center}
\vskip -0.2in
\end{figure}

Even when asymptotic complexity of two operations match, there might be an arbitrary high constant hidden in the big-O notation. To show that this is not the case we run a simple experiment, where we time a simple Jax~\citep{jax2018github} implementation of both methods computing a forward pass for varied $N$ and $d$ and report the slowdown ratio of Intention over Attention (see Figure~\ref{fig:slowdown}). Note that Intention is approximately only between 1. and 4. times slower than Attention.

%%%%%%%%%%%%%%%%%%%

Another important element is numerical stability. Since we are performing matrix inversion we need to ensure that the output of $\Sigma$ is indeed invertible. To do so, we can use the Moore-Penrose pseudoinverse instead of the regular inverse.
Additionally, similarly to attention, one can add scaling to ensure the variance at initialisation is close to 1. We derive this scaling in Section L of the Appendix.

Finally, if the desired output of the module is of fixed dimensionality one can output not only predictions over queries, but also the inferred mapping itself $\fmtfunction{w}(\fmtmatrix{K},\fmtmatrix{V})$. Note that this is not true for Attention as the corresponding object is the Attention map that is of size $N \times N$.

\subsection{Variants}
Instead of using least squares support vector machines one could also consider using Linear Discriminant Analysis (LDA) or Quadratic Discriminant Analysis (QDA). With Informer, it is as simple as adjusting the $\Sigma$ function and potentially doing a simple remapping of values inside the $\fmtfunction{w}$ function. 
\begin{propositionrep}
Solutions of Ridge Regression, LS-SVM, LDA and QDA (as well as their weighted versions) for a classification problem $\fmtmatrix{X} \in \mathbb{R}^{N \times d}, \fmtmatrix{y} \in \{0,1\}^{N}$ can all be represented as
$
\fmtvector{w}^*_\textrm{m} := \Sigma^{-1}_\textrm{m} \fmtmatrix{X}' \fmtmatrix{y}_\textrm{m},
$
where $\fmtmatrix{y}_\textrm{m}$ is relabeling of $\fmtmatrix{y}$ and $\Sigma_m$ is some square matrix summarising the data.
\end{propositionrep}
\begin{appendixproof}
\begin{itemize} 
\item Ridge Regression: For $\fmtmatrix{I}$ being an identity matrix we have $$\Sigma := \fmtmatrix{X}'\fmtmatrix{X} + C \fmtmatrix{I}$$
$$\fmtmatrix{y}_m := \fmtmatrix{y}$$ which matches the functional form of a Ridge Regression.
\item Weighted Ridge Regression: Let us define $\fmtmatrix{Z}$ as a diagonal matrix, where $\fmtmatrix{Z}_{ii} := \sqrt{w_i}$ where $w_i$ is a desired $i$th sample weight in Ridge regression. Then for $$\Sigma := (\fmtmatrix{ZX}')\fmtmatrix{ZX} + C \fmtmatrix{I}$$
$$\fmtmatrix{y}_m := \fmtmatrix{Zy}$$
we obtain
$$
\Sigma = \fmtmatrix{[X'Z'ZX]}^{-1}\fmtmatrix{X'Z'Zy} = 
 \fmtmatrix{[X'WX]}^{-1}\fmtmatrix{X'Wy}
$$which matches the functional form of a Weighted Ridge Regression.

Analogous argument works for a weighted version of any of the following models thus we skip the proofs.

Note, that this mapping can be exactly represented by \inten{} module in its default form, which means that our model is capable of representing weighted solutions without any modifications.

\item LS-SVM: Let us assume that $\fmtmatrix{X}$ has a column of 1s representing the bias. Then for $\fmtmatrix{I_{d-1}}$ being an identity matrix with a 0 in the entry corresponding to the bias dimension we have $$\Sigma := \fmtmatrix{X}'\fmtmatrix{X} + C \fmtmatrix{I_{d-1}}$$
$$\fmtmatrix{y}_m := 2\fmtmatrix{y}-1$$which matches the functional form of a LS-SVM.

This means, that in practise our default Intention is equivalent to an (weighted) LS-SVM when applied to a classification context.
\item LDA: Let us assume that $N^k$ is number of samples in class $k$, and $\bar{\fmtmatrix{X}}$ is a feature-wise mean of points, then $$\Sigma := \fmtfunction{cov}(\fmtmatrix{X}) =  (\fmtmatrix{X} - \fmtmatrix{\bar{X}} )'(\fmtmatrix{X} - \fmtmatrix{\bar{X}} )$$
$$\fmtmatrix{y}_m := \tfrac{1}{N}[ N^0 \fmtmatrix{y} +  N^1 (1-\fmtmatrix{y})]$$
It is easy to see that 
\begin{equation}
\begin{aligned}
\fmtmatrix{X}'\fmtmatrix{y}_m &= \fmtmatrix{X}'\left (\tfrac{1}{N}[ N^0 \fmtmatrix{y} +  N^1 (1-\fmtmatrix{y})] \right ) \\
& = \bar{\fmtmatrix{X}^1}' - \bar{\fmtmatrix{X}^0}'
\end{aligned}
\end{equation}
and thus the full solution becomes
$$
[(\fmtmatrix{X} - \fmtmatrix{\bar{X}} )'(\fmtmatrix{X} - \fmtmatrix{\bar{X}}) ]^{-1}(\bar{\fmtmatrix{X}^1}' - \bar{\fmtmatrix{X}^0}') = \fmtvector{w}^*_\mathrm{LDA}
$$which matches the functional form of a LDA.
\item QDA: Using same notation as before, but also denoting points belonging to class $k$  as $\fmtmatrix{X}^k$, and their feature-wise means $\bar{\fmtmatrix{X}^k}$ we get $$\Sigma := \sum_{k} \fmtfunction{cov}(\fmtmatrix{X}^k) = \sum_{k}(\fmtmatrix{X^k} - \fmtmatrix{\bar{X^k}} )'(\fmtmatrix{X^k} - \fmtmatrix{\bar{X^k}} )$$
$$\fmtmatrix{y}_m := \tfrac{1}{N}[ N^0 \fmtmatrix{y} +  N^1 (1-\fmtmatrix{y})]$$
Similarly to LDA we get
$$
[\sum_k (\fmtmatrix{X}^k - \fmtmatrix{\bar{X^k}} )'(\fmtmatrix{X^k} - \fmtmatrix{\bar{X^k}}) ]^{-1}(\bar{\fmtmatrix{X}^1}' - \bar{\fmtmatrix{X}^0}') = \fmtvector{w}^*_\mathrm{QDA}
$$which matches the functional form of a QDA.
\end{itemize}
\end{appendixproof}

This means, that in practise our default Intention is equivalent to an LS-SVM when applied to a classification context.
Given this formulation we can also easily construct kernelised versions of each of these variants by Eq.~\ref{eq:kerneltrick}.

Finally, we note that with an analogous computation we can extend Intention to also support (in)equality constraints over the solution, which can be end-to-end trainable. For example, in order to incorporate an inequality constraint $\fmtmatrix{C}' \fmtvector{w}_{\fmtmatrix{C}, \fmtvector{c}}^* \geq \fmtvector{c}$ we could simply replace the unconstrained solution $\fmtvector{w}^*$ with
$
\fmtvector{w}^*_{\fmtmatrix{C},\fmtvector{c}} := \fmtvector{w}^* - \fmtmatrix{Z}\fmtmatrix{\bar{C}}(\fmtmatrix{\bar{C}}'\fmtmatrix{Z}\fmtmatrix{\bar{C}})^{-1}(\fmtmatrix{\bar{C}}'\fmtmatrix{w}^* - \fmtvector{\bar{c}}),
$
where $\fmtmatrix{Z} := (\fmtmatrix{K}'\fmtmatrix{K})^{-1}$ and $\fmtmatrix{\bar{C}}, \fmtvector{\bar{c}}$ are the submatrix (and subvector) of the violated constraints~\citep{theil1960quadratic}. In particular if no constraints are violated then $\fmtvector{w}^*_{\fmtmatrix{C},\fmtvector{c}} := \fmtvector{w}^*$.

\section{Related work}
\label{related}

So far we have shown how Intention is related to Attention and Transformers. 
To the best of our knowledge this relation itself has not been investigated in the deep learning literature and is therefore one of the important contributions of our paper.
Additionally, \inten{} allows us to draw connections between other existing methods of deep learning that we will discuss in the following.

\paragraph{Optimisation modules} \inten{} solves the optimisation problem posed by least squares regression as part of its forward pass. 
In a similar fashion OptNet~\citep{amos2017optnet} uses a differentiable optimiser to solve convex optimisation problems at inference time. 
The family of problems that OptNet can solve is broader than \inten, at the cost of longer compute time, stability issues associated with iterative optimisers and thus potentially lack of convergence.
Also closely related, input convex neural networks~\citep{amos2017input} produce outputs that are convex wrt (part of) their input and can be minimised easily. Unlike \inten{} they lack the ability to output the minima themselves.
Another model with an embedded differentiable iterative solver is introduced by~\citep{bertinetto2018meta}. This model tackles meta-learning tasks by solving SVMs, so it is similar to OptNet but more specifically tailored to one specific optimisation problem.

\paragraph{Closed form solutions to least squares}
The closed form solution to linear regression has been used before as well~\citep{gilton2019neumann,lee2019meta}. 
In these instances this computation is used as the solution to a very specific problem and lacks the unified view and the connection to other relevant models (like Attention) that we provide. 
In addition, because of this disconnect, previous approaches did not expand their models motivated by Attention as we do with $\sigma$Intention.

% Neuman nets
% Finally the closest are~\citep{lee2019meta} and.

\paragraph{Few-shot Learning}
As we will show in this paper, \inten{} modules are a good way of integrating conditional information in few-shot learning tasks (also called meta learning or in context learning). 
Few-shot learning research encompasses a very large set of areas (for a detailed overview of few-shot learning methods in deep learning we refer the reader to~\citep{huisman2021survey}).
For the purpose of this paper we divide the space of models into three broad areas. 
Interestingly, \inten{} provides a common ground to relate them to each other:
\begin{enumerate}
    \item Similarity-based methods like the aforementioned Attention models~\citep{brown2020language}, matching networks~\citep{vinyals2016matching} and others~\citep{snell2017prototypical} rely on pointwise computations. If we think of these methods as computing some distance $\langle \fmtvector{q}_i, \fmtvector{k}_j \rangle$, we can express Intention as a special case where the distance is computed as the Mahalanobis dot product $\langle \fmtvector{q}_i, \fmtvector{k}_j\rangle _{\Sigma}$ where $\Sigma=\fmtmatrix{K}'\fmtmatrix{K}$.
    \item Optimisation-based methods like MAML~\citep{finn2017model} or Leo~\citep{rusu2018meta}. Given that its inner loop solves an optimisation problem \inten{} can be compared to this type of optimisation-based methods as well.
    Instead of solving it iteratively, however, \inten{} computes its closed form and instead of optimising the parameters of the model it optimises some latent representation.
    \item Latent model-based methods like Neural Processes~\citep{garnelo2018conditional} and others~\citep{mishra2017simple, eslami2018neural}. In this context \inten{} can be seen as an algorithm with a more expressive way of aggregating the conditioning information. 
\end{enumerate}

\section{Empirical investigation}
\label{results}

We now proceed to explore the strengths of \inten-based models empirically. 
Because we want to focus on the actual characteristic computation of each model for our experiments we strip the models down to the bare minimum and remove any bells and whistles that might improve performance but obscure a fair and direct comparison between them.
As a result our goal is not to achieve SOTA but rather compare the relative performance of different approaches.
Detailed descriptions of the tasks, experimental setups and hyper parameters can be found in the Appendix.

\subsection{Few-shot Regression}
As mentioned in Section~\ref{related}, \inten{} can be used as a conditioning module and is therefore useful for tasks that require the integration of context information. In order to evaluate this we apply \inten{} to a number of few-shot learning tasks.

\begin{figure}[ht]
\begin{center}
\centerline{\includegraphics[width=\columnwidth]{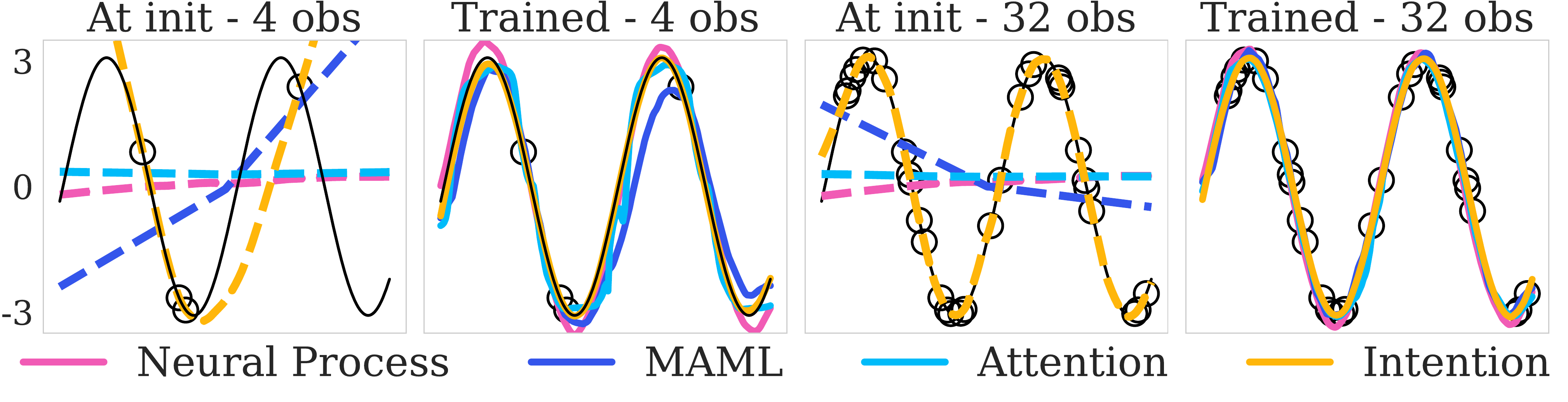}}
\caption{Sine regression experiments. Black line: ground truth, black circles: observations. The different columns show different number of context points for the same task for untrained (dashed) and trained models (solid).}
\label{fig:sine}
\end{center}
\vskip -0.2in
\end{figure}

The first set of experiments follow the few-shot regression setup described in \citep{finn2017model}, where the task is to regress sine curves with different amplitudes and shifts from a handful of observations (see Figure~\ref{fig:sine} for a visualisation of the task).
We compare performance against three common baselines for few-shot learning: gradient-based models (MAML~\citep{finn2017model}), ammortised models (Neural Processes (NPs)~\citep{garnelo2018conditional, garnelo2018neural}) and an Attention model.
All the baselines share the same fundamental architecture and hyperparameters have been tuned for best performance for each of them. 

\begin{figure}[ht]

\begin{center}
 \centerline{\includegraphics[width=\columnwidth]{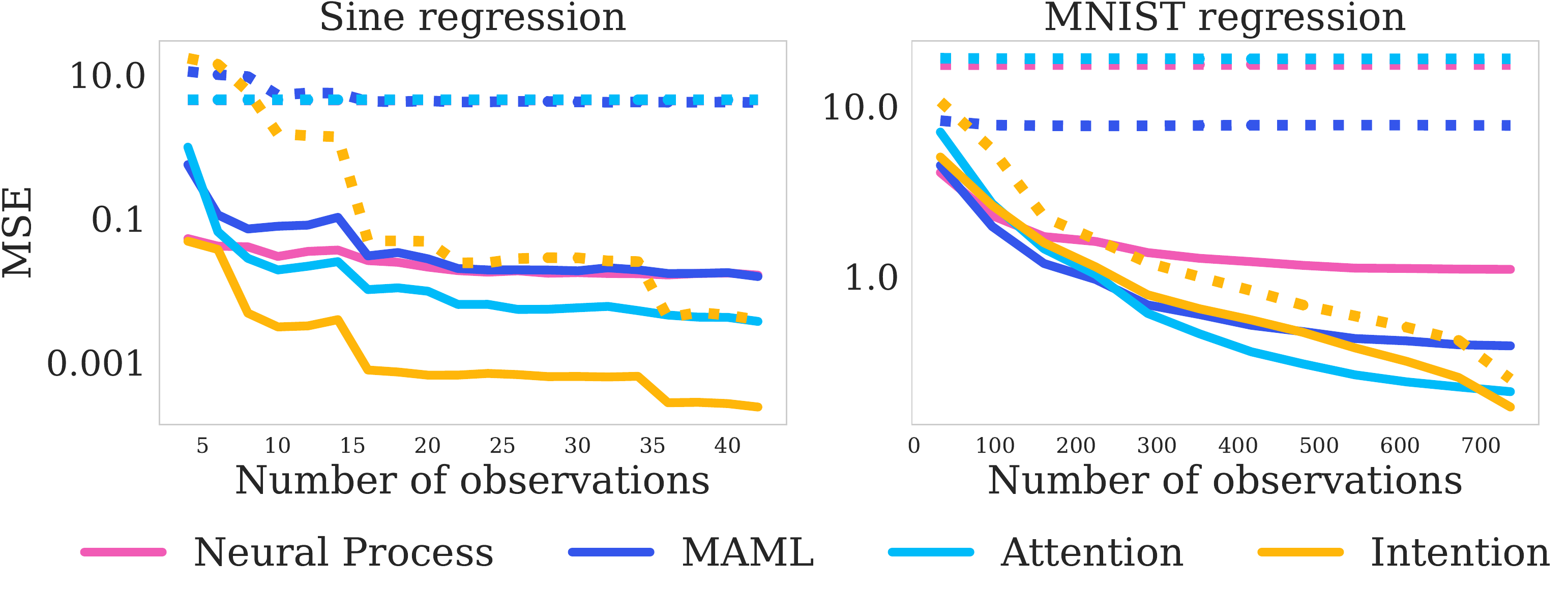}}
\caption{MSE over increasing number of observations on the two few-shot regression. A dashed line indicates the performance the model is being evaluated before training.}
\label{fig:reg_quant}
\end{center}
\vskip -0.2in
\end{figure}

We show some qualitative examples of the models' performances in Figure~\ref{fig:sine} and plot their average MSE for different numbers of observations in Figure~\ref{fig:reg_quant}.
In addition to reporting performance after training we are also plotting performance at initialisation.
The idea behind this is to show that in addition to performing well once trained \inten{} also provides a powerful inductive bias for regression.
As shown in Figure~\ref{fig:sine}, \inten{} is able to interpolate between observations and therefore achieve good performance even when untrained. With enough context points its performance matches those of trained baselines (Figure~\ref{fig:reg_quant}).
Once trained \inten{} is able to extrapolate between observations, as it has learned the common patterns of the data set. In terms of final performance \inten{} outperforms the remaining baselines on any number of observations.
     
A more complex regression task introduced in~\citep{garnelo2018conditional} is to regress the colour of individual pixels in an image, in particular of images from the MNIST data set~\citep{deng2012mnist}. 
As before, the inductive bias of \inten{} results in untrained models matching the performance of trained ones on high number of context points (Figure~\ref{fig:reg_quant}). 
The good performance of \inten{} on regression tasks even when untrained, while promising, need not come as a surprise given that the model is solving linear regression as part of its computation. 
This is a property that models in the past have taken advantage of as well~\citep{pao1994learning}. 

Once trained, \inten{} and the baselines perform similarly.
For a visualisation of the predicted images see Figure~\ref{fig:mnist} in Section~\ref{app:mnist} of the Appendix.

\begin{figure}[ht]
\begin{center}
\centerline{\includegraphics[width=\columnwidth]{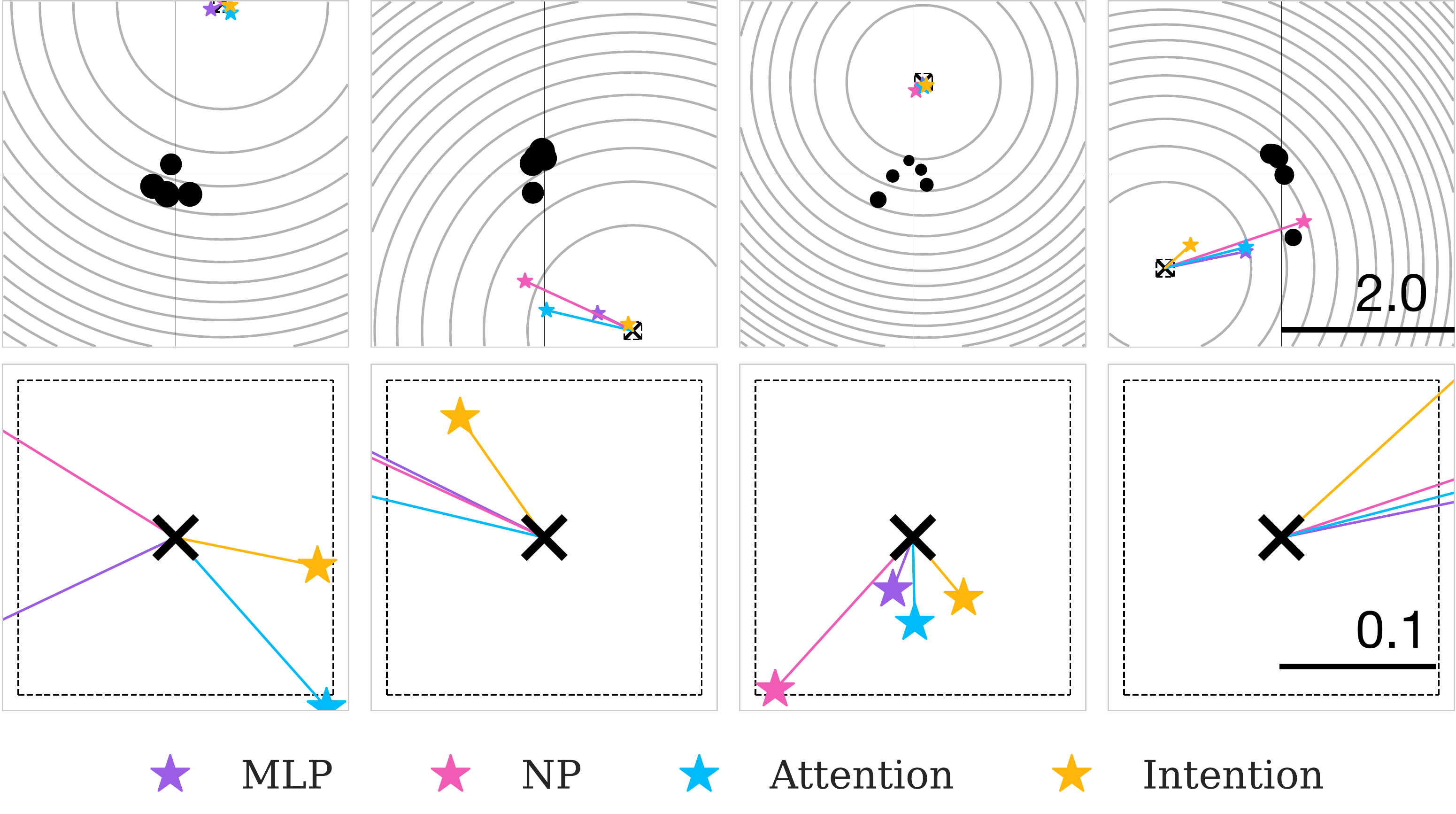}}
\caption{Policy distillation examples: the top row shows 4 examples of the experimental setup and the predictions. The reward function (a Gaussian around the target) is outlined in grey and the observations are plotted as black circles with size that is proportional to the associated reward according to the reward function. The predictions of the different models are indicated by stars. In order to better see the difference in predictions the bottom row shows a zoomed in view centered around the target. 
}
\label{fig:jaxiestjax}
\end{center}
\vskip -0.2in
\end{figure}

\subsection{Few-shot Classification}
We now move to tasks beyond regression by evaluating \inten{} on the MiniImagenet 5-shot classification task~\citep{vinyals2016matching}.
% We follow the experimental procedure of MAML and compare to it as a baseline.
For a fair comparison we implement \inten{} and all other baselines following the original architecture described in~\citep{finn2017model}.
For the implementation of \inten{} we choose the LS-SVM framework.
As shown in Table~\ref{tab:external_miniimagenet}, \inten{} outperforms all the baselines.

\begin{table}[t]
\caption{Classification accuracies for 5-shot MiniImagenet with a Matching Networks~\citep{vinyals2016matching} encoder. Mean over 1000 evaluations.}
\label{tab:external_miniimagenet}

\begin{center}
\begin{small}
\begin{sc}
\begin{tabular}{clc}
\toprule
& Model & Accuracy \\
\midrule
\color[HTML]{00BBF9}\newmoon \color{black} & Attention & 54.26 \% $\pm$ 0.51\%\\
\color[HTML]{F15BB5}\newmoon \color{black} & NP &  60.10\% $\pm$ 0.51\% \\
\color[HTML]{3455eb}\newmoon \color{black} & MAML \cite{finn2017model} & 63.15\% $\pm$ 0.91\%\\
% \informl - kernel LDA & \mg{doesnt work}\\
\color[HTML]{FFB60A}\newmoon \color{black} & \inten{} & \textbf{67.30} \% $\pm$ 0.51\%\\
\bottomrule
\end{tabular}
\end{sc}
\end{small}
\end{center}
\vskip -0.1in
\end{table}

% OLD NON-MATCHIGN NETS
\begin{table}[tb]
\caption{Classification accuracies for 5-shot MiniImagenet with a ResNet encoder. Mean over 1000 evaluations. 
}
\label{tab:internal_miniimagenet}

\begin{center}
\begin{small}
\begin{sc}
\begin{tabular}{clc}
\toprule
& Model & Accuracy \\
\midrule
% NP & 65.86 \% $\pm$ 0.52\% \\
% Attention & 63.92\% $\pm$ 0.52\%\\
&\inten{} - QDA & 66.27 \% $\pm$ 5.81\%\\
\color[HTML]{FFB60A}\newmoon \color{black} &\inten{} & 70.05 \% $\pm$ 0.51\%\\
\color[HTML]{e37405}\newmoon \color{black} &\inten{} - kernel & \textbf{70.43} \% $\pm$ 0.52\%\\
\bottomrule
\end{tabular}
\end{sc}
\end{small}
\end{center}
\vskip -0.1in
\end{table}

We further investigate how performance is affected when \inten{} is reframed as either QDA or uses a Gaussian kernel $\fmtspace{K}(\fmtvector{x},\fmtvector{y}) := \exp(\gamma \| \fmtvector{x}-\fmtvector{y}\|^2 )$. 
The results for this are shown in Table~\ref{tab:internal_miniimagenet}.
Unlike before our choice of architecture is not constrained by other baselines so we use a more powerful encoder, which results in higher overall performance. 
We observe that framing \inten{} as QDA reduces accuracy compared to LS-SVM and that the kernelised version matches the unkernelised version.
What makes the kernelised version interesting, however, is its ability to reach better performance in a smaller latent size regime (see Figure~\ref{fig:kernel_vs_not}).
This shows an interesting way of increasing model performance, that doesn't just rely on increasing model size, as is often the case for deep learning models.

\subsection{Policy Distillation}
\label{jaxiestjax}

In order to test the usefulness of \inten{} for reinforcement learning (RL) we apply it to the navigation benchmark described in \citep{finn2017model}. 
In its original version an agent is placed into an environment with a new target location for every meta iteration and observes a few trajectories before having to navigate to said target. 
This task can also be rephrased as a supervised policy distillation task by framing it as a target prediction task given some observations in the environment.
This framing, while being equivalent, decouples learning the navigation from learning the reward function and thus reduces the high variance of RL. 
This allows us to better compare models while also not have to worry about the RL algorithm part, which is not what \inten{} addresses in the first place.

\begin{figure}[ht]
\begin{center}
\centerline{\includegraphics[width=\columnwidth]{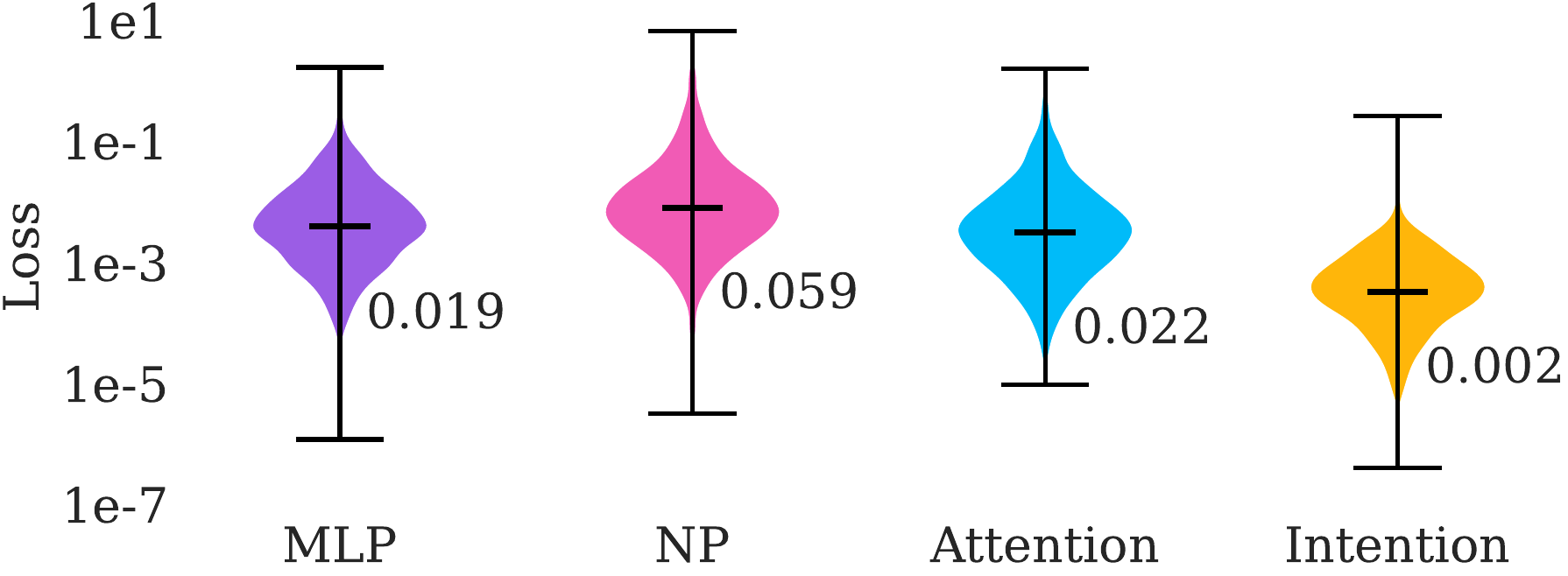}}
\caption{Quantitative policy distillation results.}
\label{fig:jaxiestjax_quant}
\end{center}
\vskip -0.2in
\end{figure}

We thus simulate the environment by first sampling a new target, then sampling $N$ locations and evaluating them according to the reward function. The goal of the model is to learn how to extract the target from these $N$ observations.

As baselines we look at different types of aggregating information in neural networks. 
From the most simple (MLP) to more sophisticated modules (Neural Processes (NPs) and Attention modules).
Figure~\ref{fig:jaxiestjax} shows some qualitative examples of the task and predictions of \inten{} and the baselines.
We also quantify the performance over 1000 trials in Figure~\ref{fig:jaxiestjax_quant} where \inten{} is the only model that shows significant improvement on the task.

% \subsection{Remapping point clouds}
\subsection{Learning generalised Kabsch algorithm}

\begin{figure}[ht]
\begin{center}
\centerline{\includegraphics[width=\columnwidth]{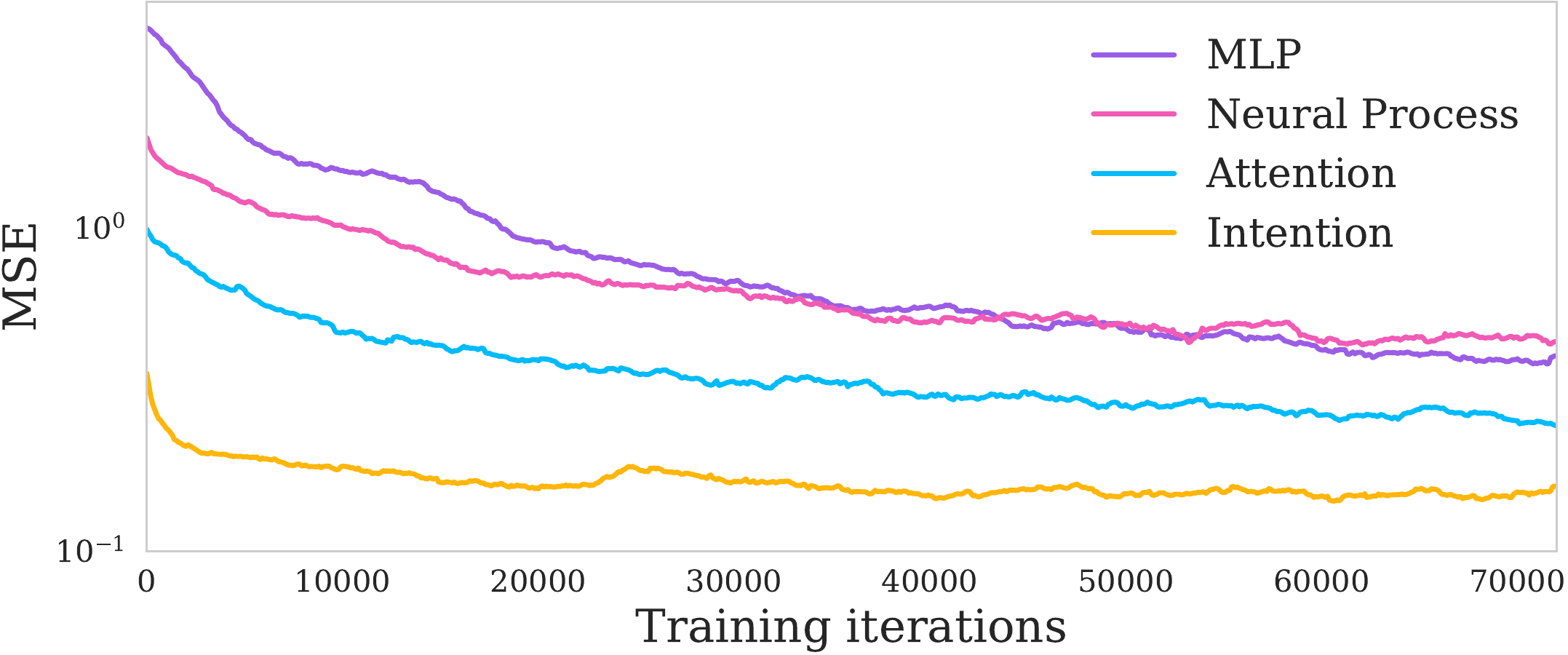}}
\caption{Quantitative results for the remapping point clouds tasks. MSE for \inten{} and the three baselines over training iterations with annotated mean loss values.}
\label{fig:jaxulino_quant}
\end{center}
\vskip -0.2in
\end{figure}
We evaluate \inten{} on a task of learning a generalised Kabsch algorithm~\cite{umeyama1991least}, where network is presented with 5 points as well as their randomly randomly rotated, translated and rescaled versions. The task is to infer this transformation and apply it to query points.

\begin{figure}[ht]

\begin{center}
\centerline{\includegraphics[width=\columnwidth]{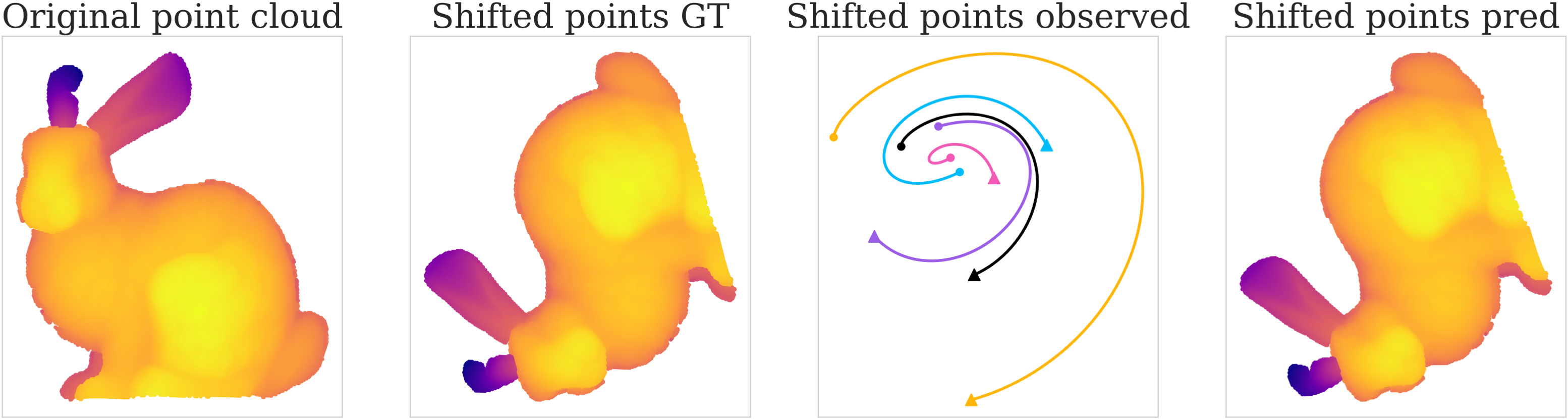}}
\caption{The remapping task evaluated on a rabbit point cloud. The first panel shows the original point cloud, the second the target obtained via the transformation shown on the 5 observations that we provide the model with in the third panel. The last plot shows the model's prediction.}
\label{fig:jaxulino}
\end{center}
\vskip -0.2in
\end{figure}

We compare the performance of \inten{} on this task to an MLP, NPs and Attention. As shown in Figure~\ref{fig:jaxulino_quant} \inten{} beats the baselines by a wide margin.
Since the learned mapping can be applied to any set of points, we task a trained \inten{} with transforming a point cloud of a rabbit, after inferring it from 5 observed points (Figure~\ref{fig:jaxulino}).

\subsection{Stacking \inten: Informers}
\label{informer}

In this final experiment we provide a proof of concept of how to use \inten{} as a universal computational block. We build a so-called Informer module by replacing the Self-Attention of a Transformer with Self-Intention as described in Section~\ref{implementing}. 
We evaluate this model on an anomaly detection task. The model is presented with 10 images from the CIFAR100 data set, 9 of which belong to the same class and 1 that is selected from a different one (outlier). 
The task is to identify the outlier. 

Table~\ref{tab:external_cifar} shows our results comparing five models. Our Transformer baseline corresponds to the model described in ~\citep{vaswani2017attention}. 
Lin-Transformer is a Transformer without the softmax computations in the Attention modules, as described in Theorem~\ref{linatt} and analogously $\sigma$-Informer is implemented as in Theorem~\ref{sigmaint}. 
NP-Former is a Transformer model with the Attention module replaced by a Neural Process module.
% Both NP-Former and MLP-Former use the Transformer like structure where we simply replaced the Attention module with NP or MLP respectively.
The results suggest a number of things: firstly, all five models benefit from stacking multiple layers and creating a deeper architecture. Secondly, for this specific task it is beneficial to incorporate bounded computations by using softmaxes both in the case of Transformers and Informers. Finally, we see a slight performance gain when using Informer over Transformer architectures. 

\begin{table}[htb]
\caption{Anomaly detection test accuracy on CIFAR100 using ResNet34 embeddings. }
\label{tab:external_cifar}
\begin{center}
\begin{small}
\begin{sc}
\begin{tabular}{clcc}
\toprule
& Model &  1 Layer & 4 Layers  \\
\midrule
\color[HTML]{F15BB5}\newmoon \color{black} & NP-Former & 93.9\% & 94.0\%  \\
% \color[HTML]{3455eb}\newmoon \color{black} & MLP-Former  &- & -\\
\color[HTML]{00BBF9}\newmoon \color{black} & Lin-Transformer  & 94.4\% & 94.4\% \\
\color[HTML]{FFB60A}\newmoon \color{black} & \infor & 94.1\% & 95.1\% \\
\color[HTML]{00BBF9}\newmoon \color{black} & Transformer  &  94.3\% & 95.3\%  \\
\color[HTML]{FFB60A}\newmoon \color{black} & $\sigma$-\infor & 95.3\% & \textbf{95.5}\% \\
\bottomrule
\end{tabular}
\end{sc}
\end{small}
\end{center}
\vskip -0.1in
\end{table}

\section{Discussion}
\label{discussion}
In this paper we explored the KVQ Space and in particular Intention. We proved that Intention can not only be seen as a generalisation of Attention but is also closely related to and serves as a unifying link for a number of other methods, such as MAML and Neural Processes. 
Motivated by its links to Attention we introduce a variation we call $\sigma$Intention, that can also be useful for some classical machine learning tasks like few-shot learning and outlier detection.

It is worth emphasizing that Intention is not meant as a replacement to Attention, but rather as an expansion of the deep learning toolbox. 
% There are also limitations to \inten. Its computation scales cubically with number of context points. 
% As we have shown, there are ways of mitigating this to some extent by calculating the covariance over latent dimensions and using kernels, but overall the compute cost remains worse than Attention.

Finally, while this type of module is very well suited to larger language tasks, given the initial exploratory nature of this paper, applying it to such tasks is unfortunately out of scope. However, given these initial results, the applicability of Intention and Informers looks promising.

\section{Acknowledgements}
We would like to thank Irene, Teresa and Manuel for their eternal patience, Pebbles for its unshakeable enthusiasm, the Foxes for their distant support, Ketjow for nothing, Sorin for making us look good and Matt for all the dancing.

%%%%%%%%%%%%%%%%%%%%%%%%%%%%%%%%%%%%%%%%%%%%%%%%%%%%%%%%%%%%%%%%%%%%%%%%%%%%%%%
%%%%%%%%%%%%%%%%%%%%%%%%%%%%%%%%%%%%%%%%%%%%%%%%%%%%%%%%%%%%%%%%%%%%%%%%%%%%%%%
% APPENDIX
%%%%%%%%%%%%%%%%%%%%%%%%%%%%%%%%%%%%%%%%%%%%%%%%%%%%%%%%%%%%%%%%%%%%%%%%%%%%%%%
%%%%%%%%%%%%%%%%%%%%%%%%%%%%%%%%%%%%%%%%%%%%%%%%%%%%%%%%%%%%%%%%%%%%%%%%%%%%%%%
% \newpage

% \onecolumn
% \begin{toappendix}

\bibliography{example_paper}
\bibliographystyle{icml2022}

\newpage
\appendix
\title{Appendix}
\date{}
\maketitle{}

\renewcommand{\thesection}{\arabic{section}}

\section{Additional model diagrams}
\begin{figure*}
\begin{center}
\centerline{\includegraphics[width=\textwidth]{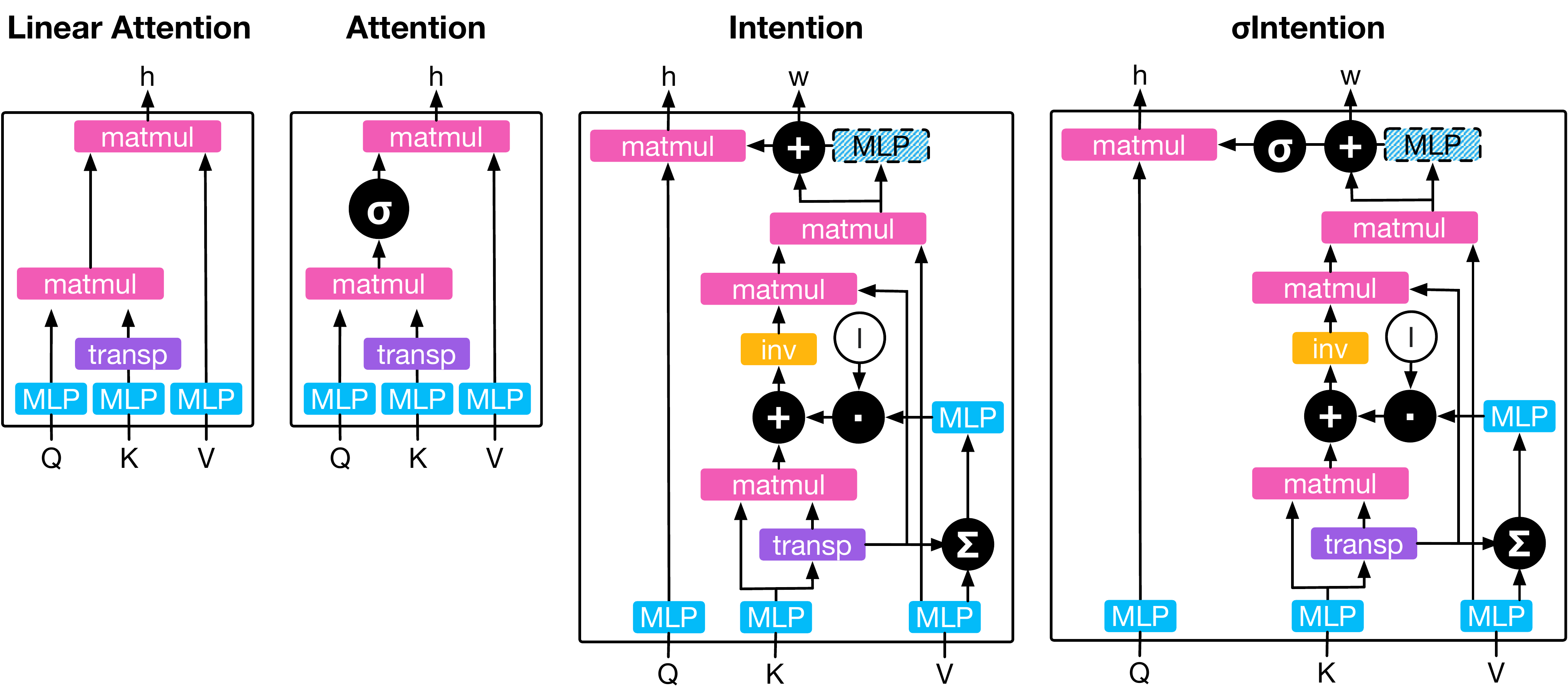}}
\caption{Diagrams of the main four models discussed in this paper and their computations. From left to right: Linear Attention, Attention and our two models which we coined~\citep{coining} Intention and $\sigma$Intenion our first moa dashed edge line indicates an optional path, `inv' stands for matrix inverse, `transp' stands for matrix transpose.}
\label{fig:all_diagrams}
\end{center}
\vskip -0.2in
\end{figure*}

In Figure~\ref{fig:all_diagrams} we show diagrams for Linear Attention, Attention, Intention and $\sigma$Intention. The Intention module varies from the more simplified one in Figure~\ref{fig:atchs} in that it also considers a learnable $\alpha$ which is a function of $\fmtmatrix{K}$ and $\fmtmatrix{V}$. This path is optional and can be replaced by a constant $\alpha$.
\label{app:diagram}

\section{Scaling experiment}
\paragraph{Data generation} The data for this experiment is generated on the fly as follows:

\begin{equation}
\begin{split}
    \fmtmatrix{W}  &\sim \mathcal{N}(0, I)\\
    \fmtmatrix{x}  &\sim \mathcal{N}(0, I)\\
    \fmtmatrix{y} & = \fmtmatrix{x}\fmtmatrix{W}
\end{split}
\end{equation}

where $\fmtmatrix{W} \in \mathbb{R}^{d \times 1}$, $\fmtmatrix{x} \in \mathbb{R}^{10 \times d}$ and $\fmtmatrix{y} \in \mathbb{R}^{10}$. $d$ is the dimension of $\fmtmatrix{x}$ which we increase as part of the experiment from 2 to 10.

\paragraph{Models} 
For each data dimension we start by setting the latent size of all layers to $s=2$.
All four models have been trained for 100000 iterations with a batch size of 64 using the MSE loss between the targets $\fmtmatrix{P}$ and predictions $\fmtmatrix{\hat{P}}$.
If the trained model does not reach the maximum error we double the size of the layer and train again $s_{\text{new}}=s^2$.
We repeat this until the model reaches the desired maximum error or until $s=8196$.
To train the model we used Adam optimiser and a learning rate of $3e^{-4}$. The rest of hyperparameters are optimised for each individual model over the same set of options and go as follows:

\begin{itemize}
    \item \textbf{Intention}:
    \begin{equation}
        \begin{aligned}
        \fmtmatrix{E}_\fmtmatrix{K} &:= \fmtfunction{e}_\fmtmatrix{K}(\fmtmatrix{K}, \fmtmatrix{V}, \fmtmatrix{Q}) \\
        \fmtmatrix{E}_\fmtmatrix{V} &:= \fmtfunction{e}_\fmtmatrix{V}(\fmtmatrix{K}, \fmtmatrix{V}, \fmtmatrix{Q}) \\
        \fmtmatrix{E}_\fmtmatrix{Q} &:= 0\\
        \fmtmatrix{w} &:= \fmtfunction{h}_\mathrm{int}( \fmtmatrix{E}_\fmtmatrix{K}, \fmtmatrix{E}_\fmtmatrix{V}, 
        \fmtmatrix{E}_\fmtmatrix{Q}) \\
        \fmtmatrix{\hat{P}}&:= \fmtfunction{e}_\fmtmatrix{P}(\fmtmatrix{w}) \\
        \end{aligned}
        \end{equation}

where $\fmtfunction{e}_\fmtmatrix{K}$, $\fmtfunction{e}_\fmtmatrix{V}$  and $\fmtfunction{e}_\fmtmatrix{P}$ are two-layer MLPs with layer sizes that change for the different input dimensions as shown in Figure\ref{fig:scaling} and ReLU non-linearities.  $\fmtfunction{h}_\mathrm{int}$ is an Intention  module as described in the main paper. 
We set $\fmtmatrix{E}_\fmtmatrix{Q}$ to 0 because we don't need any queries in the Intention computations as we are outputting $\fmtmatrix{w}$ and not $\fmtmatrix{P}$.

\item \textbf{Attention}:
    \begin{equation}
        \begin{aligned}
        \fmtmatrix{E}_\fmtmatrix{K} &:= \fmtfunction{e}_\fmtmatrix{K}(\fmtmatrix{K}, \fmtmatrix{V}, \fmtmatrix{Q}) \\
        \fmtmatrix{S}_\fmtmatrix{K} &:= \max(0,\fmtmatrix{E}_\fmtmatrix{K}) + \tau \times \min(0,\fmtmatrix{E}_\fmtmatrix{K})\\
        \fmtmatrix{E}_\fmtmatrix{Q} &:= \fmtfunction{e}_\fmtmatrix{Q}(\fmtmatrix{K}, \fmtmatrix{V}, \fmtmatrix{Q}) \\
        \fmtmatrix{S}_\fmtmatrix{Q} &:= \max(0,\fmtmatrix{E}_\fmtmatrix{Q}) + \tau \times \min(0,\fmtmatrix{E}_\fmtmatrix{Q})\\
        \fmtmatrix{E}_\fmtmatrix{V} &:= \fmtfunction{e}_\fmtmatrix{V}(\fmtmatrix{K}, \fmtmatrix{V}, \fmtmatrix{Q}) \\
        \fmtmatrix{S}_\fmtmatrix{V} &:= \max(0,\fmtmatrix{E}_\fmtmatrix{V}) + \tau \times \min(0,\fmtmatrix{E}_\fmtmatrix{V})\\
        \fmtmatrix{E}_\fmtmatrix{H} &:= \fmtfunction{h}_\mathrm{MHA}( \fmtmatrix{S}_\fmtmatrix{K}, \fmtmatrix{S}_\fmtmatrix{V}, \fmtmatrix{S}_\fmtmatrix{Q}) \\
        \fmtmatrix{S}_\fmtmatrix{H} &:= \max(0,\fmtmatrix{E}_\fmtmatrix{H}) + \tau \times \min(0,\fmtmatrix{E}_\fmtmatrix{H})\\
        \fmtmatrix{\hat{P}}&:= \fmtfunction{e}_\fmtmatrix{P}(\fmtmatrix{S}_\fmtmatrix{H}) \\
        \end{aligned}
        \end{equation}

where $\fmtfunction{e}_\fmtmatrix{K}$,  $\fmtfunction{e}_\fmtmatrix{V}$ and $\fmtfunction{e}_\fmtmatrix{Q}$ are one-layer MLPs with varying sizes layer sizes. 
$\fmtfunction{h}_\mathrm{MHA}$ is a Multihead Attention module with 5 heads. $\fmtfunction{e}_\fmtmatrix{P}$ is a one-layer decoding MLP. 
$\tau$ is the negative slope of the leaky ReLU with a value of 0.01.

\item \textbf{Neural Process}:
    \begin{equation}
        \begin{aligned}
        \fmtmatrix{E}_\fmtmatrix{H} &:= \fmtfunction{e}_\fmtmatrix{H}(\fmtmatrix{K}, \fmtmatrix{V}, \fmtmatrix{Q}) \\
        \fmtmatrix{\hat{P}} &:= \fmtfunction{e}_\fmtmatrix{P}(\tfrac{1}{N}\sum_i\fmtmatrix{E}_{\fmtmatrix{H}_i}) \\
        \end{aligned}
        \end{equation}

where $\fmtfunction{e}_\fmtmatrix{H}$ is an two-layer MLP with varying layer sizes and ReLU non-linearities. $\fmtfunction{e}_\fmtmatrix{P}$ is a four-layer MLP with varying layer sizes and ReLU non-linearities. 

\item \textbf{MLP}:

   \begin{equation}
        \begin{aligned}
        \fmtmatrix{E}_\fmtmatrix{\hat{P}} &:= \fmtfunction{e}_\fmtmatrix{P}(\fmtmatrix{K}, \fmtmatrix{V}, \fmtmatrix{Q}) \\
        \end{aligned}
        \end{equation}
        
where $\fmtfunction{e}_\fmtmatrix{P}$ is an four layer MLP with varying layer sizes and ReLU non-linearities.

\end{itemize}

\section{Comparing the computations of Attention, Linear Attention, Intention and $\sigma$Intention}
\label{app:jaxu}

\paragraph{Data generation} To create the plot in Figure~\ref{fig:attvsint} we generate the following data:

% \begin{equation}
% \begin{split}
$$
    \fmtmatrix{w} \sim          
    \mathcal{N}\bigg(\begin{bmatrix}
          0 \\
          0 \\
         \end{bmatrix}, \begin{bmatrix}
          10, 0 \\
          0, 10 \\
         \end{bmatrix}\bigg)
$$
    $$\fmtmatrix{K} \sim \mathcal{N}\bigg(\begin{bmatrix}
      0 \\
      0 \\
     \end{bmatrix}, \begin{bmatrix}
      0.7, 0.9 \\
      0.9, 1.0 \\
     \end{bmatrix}\bigg)
     $$
    $$\fmtmatrix{Q}_\text{in}  \sim \mathcal{U}(-1, 1)$$ 
    $$\fmtmatrix{Q}_\text{ex}  \sim \mathcal{U}(-25, 25)$$
    $$\fmtmatrix{V} = \fmtmatrix{K}' \fmtmatrix{w}$$
    $$\fmtmatrix{P}_\text{in}  = \fmtmatrix{Q}_\text{in}'\fmtmatrix{w}$$
    $$\fmtmatrix{P}_\text{ex}  = \fmtmatrix{Q}_\text{ex}'\fmtmatrix{w}$$
% \end{split}
% \end{equation}

\paragraph{Models}
We compute the predictions $\fmtmatrix{\hat{P}}$ for all the queries both in the interpolation experiment $\fmtmatrix{Q}_{\text{in}}$ and extrapolation experiment $\fmtmatrix{Q}_{\text{ex}}$ as follows:
\begin{itemize}
    \item Attention: $\fmtmatrix{\hat{P}} = \sigma(\fmtmatrix{Q}\fmtmatrix{K}')\fmtmatrix{V}$
    \item Linear Attention: $\fmtmatrix{\hat{P}} = \fmtmatrix{Q}\fmtmatrix{K}'\fmtmatrix{V}$
    \item $\sigma$Intention: $\fmtmatrix{\hat{P}} = \sigma(\fmtmatrix{Q}(\fmtmatrix{K}'\fmtmatrix{K})^{-1}\fmtmatrix{K}')\fmtmatrix{V}$
    \item Intention: $\fmtmatrix{\hat{P}} = \fmtmatrix{Q}(\fmtmatrix{K}'\fmtmatrix{K})^{-1}\fmtmatrix{K}'\fmtmatrix{V}$
\end{itemize}

where $\sigma$ is the softmax operation.

\section{Few-shot regression}
\subsection{Sine regression}
\label{app:sine}
\paragraph{Data generation} We follow the experimental setup of~\citep{finn2017model}. The data for this experiment is generated on the fly as follows:

\begin{equation}
\begin{split}
    b &\sim \mathcal{U}(0, \pi)\\
    a& \sim \mathcal{U}(0.1, 5)\\
    y &= a \times \sin (x - b)
\end{split}
\end{equation}

$\forall x \in [-6, 6)$. We use $M=200$ points as queries and targets and $N=10$ of those points as keys and values.
The input to our models is therefore: keys $\fmtmatrix{K} \in \mathbb{R}^{10 \times 1}$,  values $\fmtmatrix{V} \in \mathbb{R}^{10 \times 1}$, queries $\fmtmatrix{Q} \in \mathbb{R}^{200 \times 1}$ and targets for the predictions $\fmtmatrix{P} \in \mathbb{R}^{200 \times 1}$.

\paragraph{Models} All four models have been trained for 50000 iterations with a batch size of 8 using the MSE loss between the targets $\fmtmatrix{P}$ and predictions $\fmtmatrix{\hat{P}}$. The rest of hyperparameters are optimised for each individual model over the same set of options and go as follows:
\begin{itemize}
    \item \textbf{Intention}:
    \begin{equation}
        \begin{aligned}
        \fmtmatrix{E}_\fmtmatrix{K} &:= \fmtfunction{e}_\fmtmatrix{K}(\fmtmatrix{K}) \\
        \fmtmatrix{E}_\fmtmatrix{Q} &:= \fmtfunction{e}_\fmtmatrix{K}(\fmtmatrix{Q}) \\
        \fmtmatrix{E}_\fmtmatrix{V} &:= \fmtfunction{e}_\fmtmatrix{K}(\fmtmatrix{V}) \\
        \fmtmatrix{\hat{P}} &:= \fmtfunction{h}_\mathrm{int}( \fmtmatrix{E}_\fmtmatrix{K}, \fmtmatrix{E}_\fmtmatrix{V}, \fmtmatrix{E}_\fmtmatrix{Q}) \\
        \end{aligned}
        \end{equation}

where $\fmtfunction{e}_\fmtmatrix{K}$ is an encoding MLP with layer sizes [1000, 1000, 1000, 1000] and ReLU non-linearities.  $\fmtfunction{e}_\fmtmatrix{K}$ is shared across keys, values and queries. $\fmtfunction{h}_\mathrm{int}$ is an Intention  module as described in the main paper. To train the model we used Adam optimiser and a learning rate of $3e^{-4}$.

\item \textbf{Attention}:
    \begin{equation}
        \begin{aligned}
        \fmtmatrix{E}_\fmtmatrix{K} &:= \fmtfunction{e}_\fmtmatrix{K}(\fmtmatrix{K}) \\
        \fmtmatrix{E}_\fmtmatrix{Q} &:= \fmtfunction{e}_\fmtmatrix{Q}(\fmtmatrix{Q}) \\
        \fmtmatrix{E}_\fmtmatrix{V} &:= \fmtfunction{e}_\fmtmatrix{V}(\fmtmatrix{V}) \\
        \fmtmatrix{E}_\fmtmatrix{H} &:= \fmtfunction{h}_\mathrm{MHA}( \fmtmatrix{E}_\fmtmatrix{K}, \fmtmatrix{E}_\fmtmatrix{V}, \fmtmatrix{E}_\fmtmatrix{Q}) \\
        \fmtmatrix{\hat{P}}&:= \fmtfunction{e}_\fmtmatrix{P}(\fmtmatrix{E}_\fmtmatrix{H}) \\
        \end{aligned}
        \end{equation}

where both $\fmtfunction{e}_\fmtmatrix{K}$ and $\fmtfunction{e}_\fmtmatrix{V}$ are encoding MLPs with layer sizes [128, 128, 128] and ReLU non-linearities.  $\fmtfunction{e}_\fmtmatrix{K}$ is shared across keys and queries. $\fmtfunction{h}_\mathrm{MHA}$ is a Multihead Attention module with 5 heads. $\fmtfunction{e}_\fmtmatrix{P}$ is a decoding MLP with layer sizes [128, 128, 1] and ReLU non-linearities.  To train the model we used Adam optimiser and a learning rate of $1e^{-4}$.

\item \textbf{Neural Process}:
    \begin{equation}
        \begin{aligned}
        \fmtmatrix{E}_\fmtmatrix{H} &:= \fmtfunction{e}_\fmtmatrix{H}(\fmtmatrix{K}, \fmtmatrix{V}) \\
        \fmtmatrix{\hat{P}} &:= \fmtfunction{e}_\fmtmatrix{P}(\tfrac{1}{N}\sum_i\fmtmatrix{E}_{\fmtmatrix{H}_i}, \fmtmatrix{Q}) \\
        \end{aligned}
        \end{equation}

where $\fmtfunction{e}_\fmtmatrix{H}$ is an encoding MLP with layer sizes [2000, 2000, 16] and ReLU non-linearities. $\fmtfunction{e}_\fmtmatrix{P}$ is a decoding MLP with layer sizes [2000, 2000, 2000 1] and ReLU non-linearities.  To train the model we used Adam optimiser and a learning rate of $1e^{-4}$.

\item \textbf{MAML}:

    \begin{equation}
    \begin{aligned}
    \fmtmatrix{\hat{V}} &:= \fmtfunction{e}_\fmtmatrix{P}(\fmtmatrix{K}) \\
    \fmtmatrix{\hat{P}} &:= \fmtfunction{e'}_\fmtmatrix{P}(\fmtmatrix{Q}) \\
    \end{aligned}
    \end{equation}
        
where $\fmtfunction{e}_\fmtmatrix{P}$ is an encoding MLP with layer sizes [128, 128, 128, 128, 1] and ReLU non-linearities and $\fmtfunction{e'}_\fmtmatrix{P}$ is the same MLP after being updated using 3 iterations of gradient descent on the loss $\mathrm{MSE}(\fmtmatrix{\hat{V}}, \fmtmatrix{V})$ with an inner learning rate of 0.1. To train the model we used Adam optimiser and an outer learning rate of $3e^{-3}$.

\end{itemize}

\subsection{MNIST}
\label{app:mnist}
\paragraph{Data} At each iteration we sample one of the images from the MNIST data set~\citep{deng2012mnist}. We sample 264 pixels at random and pass their position in the image as keys $\fmtmatrix{P}$, the colour of each of them as values $\fmtmatrix{V}$ and the positions of all the pixels in the image as queries $\fmtmatrix{Q}$. 

\paragraph{Models} For the MNIST regression experiments we use the same model architecture as for the sine curves experiments but add positional encoding to the pixel positions.

\begin{figure}[ht]

\begin{center}
\centerline{\includegraphics[width=\columnwidth]{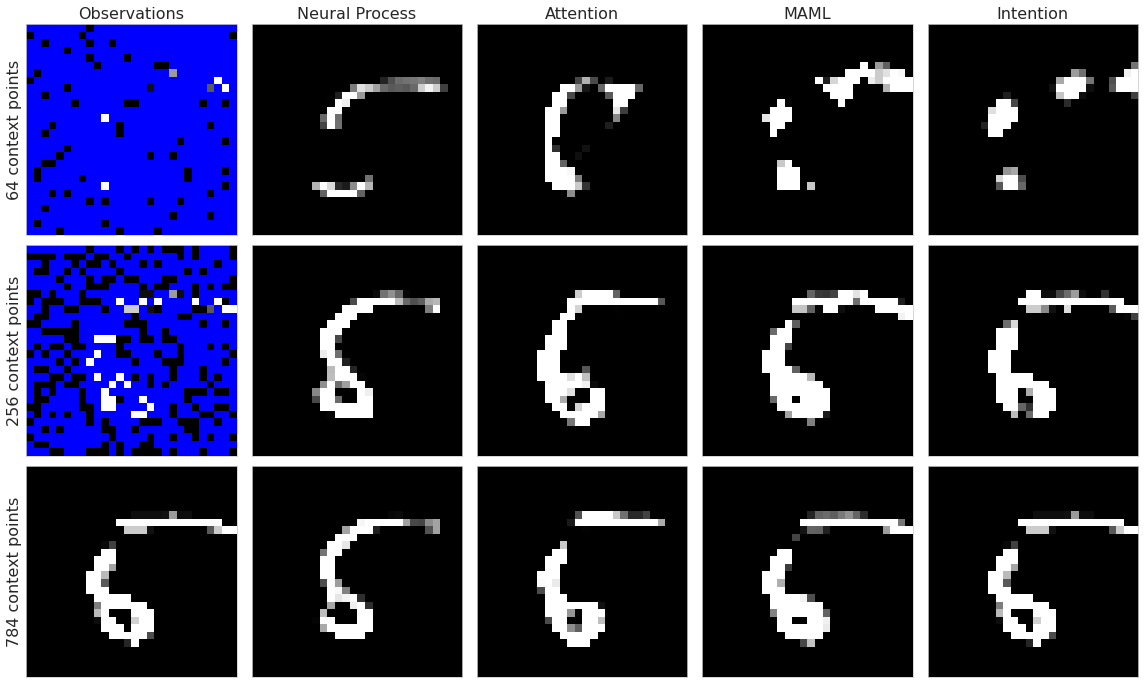}}
\caption{Image completion as few-shot regression.}
\label{fig:mnist}
\end{center}
\vskip -0.2in
\end{figure}

\paragraph{Additional results} In addition to the quantitative results we can also plot the images regressed by the different baselines as shown in Figure~\ref{fig:mnist}.

\section{Few-shot Classification}
\label{app:miniimagenet}

\paragraph{Data set} We use the MiniImagenet data set introduced in~\citep{vinyals2016matching}. It consists of 100 classes (80 train, 10 validation and 10 test) from the ImageNet data set that have been scaled down to $28 \times 28$. Each class has 600 images. For our experiments we randomly select 5 classes and pick 5 images from each class as context and 15 as targets.

\paragraph{Training} We train Intention using an MSE loss and Neural Processes and Attention using a softmax-crossentropy loss. 
We train to convergence using stochastic gradient descent with momentum (with a value of 0.9) and additive weight decay (decay rate of $5e^{-4}$).
We apply an exponential scheduling function to our learning rate (the hyperparameters of this scheduling for each model is outlined below). 

\paragraph{Models} 
\begin{itemize}
    \item \textbf{Intention}:
    
    \begin{equation}
        \begin{aligned}
        \fmtmatrix{E}_\fmtmatrix{K} &:= \phi_\mathcal{K}(\fmtfunction{e}_\fmtmatrix{H}(\fmtfunction{e}_\fmtmatrix{enc}(\fmtmatrix{K}))) \\
        \fmtmatrix{E}_\fmtmatrix{Q} &:= \phi_\mathcal{K}(\fmtfunction{e}_\fmtmatrix{H}(\fmtfunction{e}_\fmtmatrix{enc}(\fmtmatrix{Q}) )) \\
        \fmtmatrix{E}_\fmtmatrix{V} &:=  \fmtmatrix{V}\\
        \fmtmatrix{\hat{P}} &:= \fmtmatrix{E}_\fmtmatrix{Q} \Sigma_m^{-1} \fmtmatrix{E}'_\fmtmatrix{K}\fmtmatrix{y}_m \\
        \end{aligned}
    \end{equation}

where $\fmtfunction{e}_\fmtmatrix{enc}$ is the encoder, which is one of either a ResNet12 architecture~\citep{he2016deep} or the encoder used the Matching Nets~\citep{vinyals2016matching}.  $\fmtfunction{e}_\fmtmatrix{H}$ is a one-layer MLP with latent size 8192 for the non-kernelised Intention and 32 for the kernelised version.
Both $\fmtfunction{e}_\fmtmatrix{enc}$ and $\fmtfunction{e}_\fmtmatrix{H}$ are shared by $\fmtmatrix{K}$ and $\fmtmatrix{Q}$.
$\phi_\mathcal{K}(\fmtmatrix{X})$ is a kernel function and for our experiments we implement both a linear as well as a Gaussian kernel. 
$\Sigma_m$ and $\fmtmatrix{y}_m$ correspond to some function that estimates the covariance and to some function that processes the targets respectively.
More details on how these are implemented for LS-SVM and for QDA can be found in Section B of the Appendix.

The learning rate schedule parameters are: initial learning rate=0.1, decay rate 0.8, start of schedule at 10000 steps and for 10000 steps.

\item \textbf{Attention}:
    \begin{equation}
        \begin{aligned}
        \fmtmatrix{E}_\fmtmatrix{K} &:= \fmtfunction{e}_\fmtmatrix{H}(\fmtfunction{e}_\fmtmatrix{enc}(\fmtmatrix{K})) \\
        \fmtmatrix{E}_\fmtmatrix{Q} &:= \fmtfunction{e}_\fmtmatrix{H}(\fmtfunction{e}_\fmtmatrix{enc}(\fmtmatrix{Q})) \\
        \fmtmatrix{E}_\fmtmatrix{V} &:= \fmtfunction{e}_\fmtmatrix{H}(\fmtfunction{e}_\fmtmatrix{V}(\fmtmatrix{V}) ) \\
        \fmtmatrix{E}_\fmtmatrix{H} &:= \fmtfunction{h}_\mathrm{MHA}( \fmtmatrix{E}_\fmtmatrix{K}, \fmtmatrix{E}_\fmtmatrix{V}, \fmtmatrix{E}_\fmtmatrix{Q}) \\
        \fmtmatrix{\hat{P}}&:=\sigma( \fmtfunction{e}_\fmtmatrix{P}(\fmtmatrix{E}_\fmtmatrix{H})) \\
        \end{aligned}
        \end{equation}

where $\fmtfunction{e}_\fmtmatrix{enc}$ is the same encoder as used in~\citep{vinyals2016matching} and $\fmtfunction{e}_\fmtmatrix{H}$ is a one-layer MLP with layer size [512]. $\fmtfunction{e}_\fmtmatrix{V}$ is an MLP with layer sizes [512, 512, 512] and $\fmtfunction{e}_\fmtmatrix{P}$ is an MLP with layer sizes [512, 512, 512, 5] and ReLU non-linearities.
$\fmtfunction{h}_\mathrm{MHA}$ is a Multihead Attention module with 5 heads. 
The learning rate schedule parameters are: initial learning rate=0.05, decay rate 0.8, start of schedule at 10000 steps and for 10000 steps.

\item \textbf{Neural Process}:
    \begin{equation}
        \begin{aligned}
        \fmtmatrix{E}_\fmtmatrix{K} &:= \fmtfunction{e}_\fmtmatrix{H}(\fmtfunction{e}_\fmtmatrix{enc}(\fmtmatrix{K})) \\
        \fmtmatrix{E}_\fmtmatrix{Q} &:= \fmtfunction{e}_\fmtmatrix{H}(\fmtfunction{e}_\fmtmatrix{enc}(\fmtmatrix{Q})) \\
        \fmtmatrix{\hat{P}} &:= \fmtfunction{e}_\fmtmatrix{P}\bigg(\Big[\tfrac{1}{N^0}\sum_i\fmtmatrix{E}^0_{\fmtmatrix{K}_i}; \dots ; \tfrac{1}{N^4}\sum_i\fmtmatrix{E}^4_{\fmtmatrix{K}_i}\Big], \fmtmatrix{E}_\fmtmatrix{Q}\bigg) \\
        \end{aligned}
        \end{equation}

where $\fmtfunction{e}_\fmtmatrix{enc}$ is the same encoder as used in~\citep{vinyals2016matching} and $\fmtfunction{e}_\fmtmatrix{H}$ is a one-layer MLP with layer size [2048]. $\fmtfunction{e}_\fmtmatrix{V}$ is an MLP with layer sizes [512, 512, 512] and $\fmtfunction{e}_\fmtmatrix{P}$ is an MLP with layer sizes [1024, 5] and ReLU non-linearities.
The superscript $k$ in $\fmtmatrix{E}^k_{\fmtmatrix{K}}$ indicates that we are selecting all of the elements of $\fmtmatrix{E}_{\fmtmatrix{K}}$ that belong to class $k$ and $N^k$ is the number of examples of class $k$.
The learning rate schedule parameters are: initial learning rate=0.1, decay rate 0.8, start of schedule at 10000 steps and for 10000 steps.

\end{itemize}

\subsection{Kernel performance experiments}
\begin{figure}[ht]
\begin{center}
\centerline{\includegraphics[width=\columnwidth]{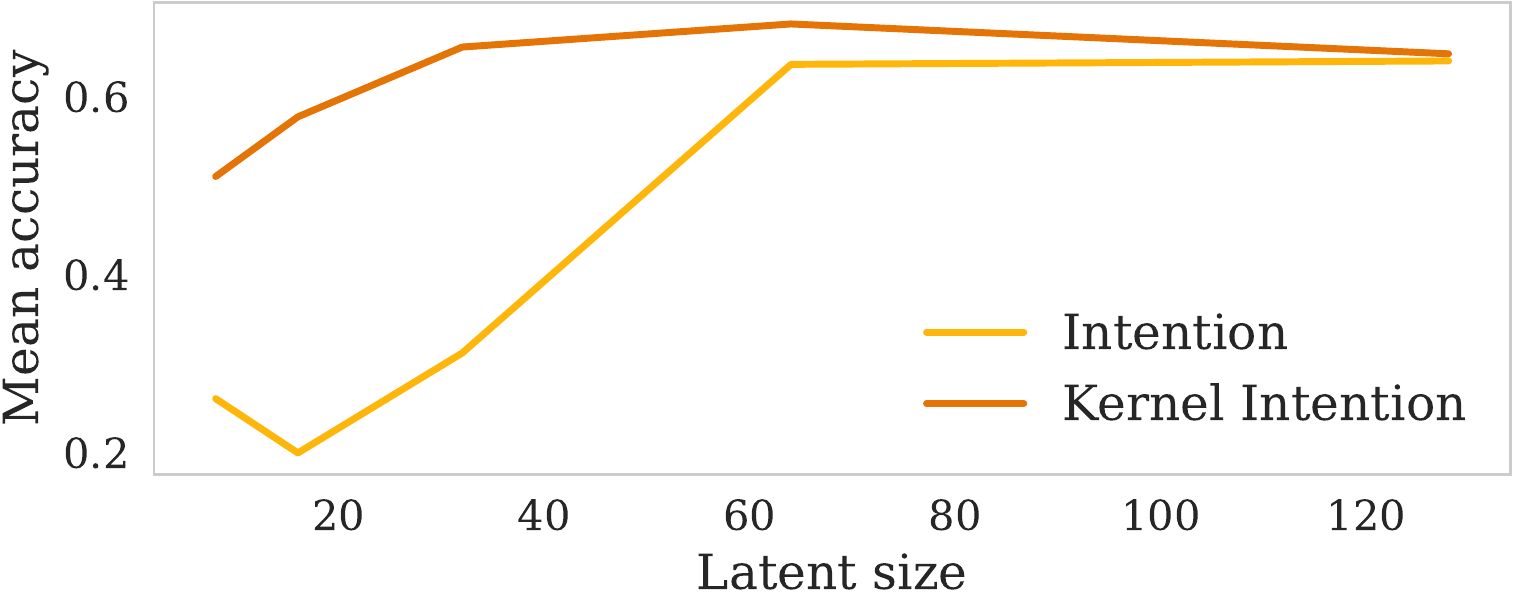}}
\caption{Accuracy as a function of model size for Intention vs its kernelised version.
% \woj{we should unify style of plots. Line styles, existance/lack of markers; legend borders etc.} 
}
\label{fig:kernel_vs_not}
\end{center}
\vskip -0.2in
\end{figure}

We compare the performance of regular Intention vs kernelised Intention using a Gaussian kernel. To do so we follow the same experimental protocol as above but vary the latent size for both models to be one of: 16, 32, 64 or 128.
We plot their performances in Figure~\ref{fig:kernel_vs_not}, where we can see that the kernelised version is able to achieve higher accuracy even with smaller latent sizes.

\section{Policy Distillation}
\label{app:pd}
\paragraph{Data} At each iteration we sample a new target location:
$$t_x,t_y \sim \mathcal{N}(0, 1)$$

We are given 5 observations from this environment by randomly sampling $x$ and $y$ coordinates and calculating the squared distance to the target: 
$$x_i, y_i \sim \mathcal{N}(0, 1/4)$$  
$$
o_i := (x_i, y_i, d_i) := (x_i, y_i, \| [x_i, y_i]  - [t_x, t_y] \|^2)
$$

The inputs to the model are contained in one matrix: 

$$\fmtmatrix{O} = \{ o_i \}_{i=1}^5  \in \mathbb{R}^{5\times 2}$$

Given $\fmtmatrix{O}$ the task is to predict $[t_x, t_y]$.

% The optimal policy is given by
% $$
% \pi^*(\{o_i\}_{i=1}^5) := \arg\min_{t'} \sum_{i=1}^5 \| \| [x_i, y_i]  - [t'_x, t'_y] \|^2 - d_i \|^2
% $$

For all models we use $\fmtmatrix{K} = \fmtmatrix{V} = \fmtmatrix{Q} := \fmtmatrix{O}$.

\paragraph{Models}
\begin{itemize}
    \item \textbf{Intention}:
    \begin{equation}
        \begin{aligned}
        \fmtmatrix{E}_\fmtmatrix{K} &:= [\fmtmatrix{K}, \fmtmatrix{V}, \fmtmatrix{Q}] \\
        \fmtmatrix{E}_\fmtmatrix{V} &:= \fmtfunction{e}_\fmtmatrix{V}(\fmtmatrix{K}, \fmtmatrix{V}, \fmtmatrix{Q}) \\
        \fmtmatrix{E}_\fmtmatrix{Q} &:= 0\\
        \fmtmatrix{w} &:= \fmtfunction{h}_\mathrm{int}( \fmtmatrix{E}_\fmtmatrix{K}, \fmtmatrix{E}_\fmtmatrix{V}, 
        \fmtmatrix{E}_\fmtmatrix{Q}) \\
        \fmtmatrix{\hat{P}}&:= \fmtfunction{e}_\fmtmatrix{P}(\fmtmatrix{w}) \\
        \end{aligned}
        \end{equation}

where $\fmtfunction{e}_\fmtmatrix{V}$ and $\fmtfunction{e}_\fmtmatrix{P}$ are two-layer MLPs with layer sizes [512, 512] and [128, 2] respectively and ReLU non-linearities.  $\fmtfunction{h}_\mathrm{int}$ is an Intention  module as described in the main paper. 
To train the model we use Adam optimiser with a learning rate of $3e-5$.

\item \textbf{Attention}:
    \begin{equation}
        \begin{aligned}
        \fmtmatrix{E}_\fmtmatrix{K} &:= [\fmtmatrix{K}, \fmtmatrix{V}, \fmtmatrix{Q}] \\
        \fmtmatrix{E}_\fmtmatrix{V} &:= \fmtfunction{e}_\fmtmatrix{V}(\fmtmatrix{K}, \fmtmatrix{V}, \fmtmatrix{Q}) \\
        \fmtmatrix{E}_\fmtmatrix{Q} &:= 
        [\fmtmatrix{K}, \fmtmatrix{V}, \fmtmatrix{Q}]\\
        \fmtmatrix{w} &:= \fmtfunction{h}_\mathrm{MHA}( \fmtmatrix{E}_\fmtmatrix{K}, \fmtmatrix{E}_\fmtmatrix{V}, 
        \fmtmatrix{E}_\fmtmatrix{Q}) \\
        \fmtmatrix{\hat{P}}&:= \fmtfunction{e}_\fmtmatrix{P}(\fmtmatrix{w}) \\
        \end{aligned}
        \end{equation}

where $\fmtfunction{e}_\fmtmatrix{V}$ and $\fmtfunction{e}_\fmtmatrix{P}$ are two-layer MLPs with layer sizes [128, 128] and [2048, 2] respectively and ReLU non-linearities.  $\fmtfunction{h}_\mathrm{int}$ is an Multihead Attention module with 5 heads. 
To train the model we use Adam optimiser with a learning rate of $3e-5$.

\item \textbf{Neural Process}:
    \begin{equation}
        \begin{aligned}
        \fmtmatrix{E}_\fmtmatrix{H} &:= \fmtfunction{e}_\fmtmatrix{H}([\fmtmatrix{K}, \fmtmatrix{V}, \fmtmatrix{Q}]) \\
         \fmtmatrix{\hat{P}} &:= \fmtfunction{e}_\fmtmatrix{P}(\tfrac{1}{N}\sum_i\fmtmatrix{E}_{\fmtmatrix{H}_i}) \\
        \end{aligned}
        \end{equation}

where $\fmtfunction{e}_\fmtmatrix{H}$ and $\fmtfunction{e}_\fmtmatrix{P}$ are two MLPs with layer sizes [512, 512] and [512, 512, 128, 2] respectively and ReLU non-linearities. 
To train the model we use Adam optimiser with a learning rate of $3e-4$.

\item \textbf{MLP}:
    \begin{equation}
        \begin{aligned}
         \fmtmatrix{\hat{P}} &:= \fmtfunction{e}_\fmtmatrix{P}([\fmtmatrix{K}, \fmtmatrix{V}, \fmtmatrix{Q}]) \\
        \end{aligned}
        \end{equation}

where $\fmtfunction{e}_\fmtmatrix{P}$ is an MLP with layer sizes [512, 512, 512, 2] and ReLU non-linearity. 
To train the model we use Adam optimiser with a learning rate of $3e-4$.

\end{itemize}

%%%%%%%%%%%%%%%%%%%%%%%%%%%%%%%%%%%%%%%%%%%%%%%%%%%%%%%%%%%%%%%%%%%%%%%%%%%%%%%
%%%%%%%%%%%%%%%%%%%%%%%%%%%%%%%%%%%%%%%%%%%%%%%%%%%%%%%%%%%%%%%%%%%%%%%%%%%%%%%

\section{Learning generalised Kabsch algorithm}
\paragraph{Task} We are given a set of $2N+M$ points: N points $\fmtvector{k}_i \in \mathbb{R}^2$, N points $\fmtvector{v}_i \in \mathbb{R}^2$ and $M$ points $\fmtvector{q}_j \in \mathbb{R}^2$. 
These points can be mapped with the following function:
$f_\theta(\fmtvector{k}_i) = a_\theta \cdot \fmtmatrix{R}_\theta \fmtvector{k}_i + \fmtvector{w}_\theta + \epsilon =: \fmtvector{v}_i$
where $\fmtvector{w}_\theta$ is a translation vector, $\fmtmatrix{R}_\theta$ is a rotation matrix, $a_\theta \in \mathbb{R}$ is a re-scaling factor and $\epsilon$ a 2-dimensional noise vector.
The goal of the network is to learn to extract $\mathbb{E}[f](\fmtvector{k}) = a_\theta \cdot \fmtmatrix{R}_\theta \fmtvector{k} + \fmtvector{w}_\theta$
and apply it to $\fmtvector{q}_i$ to produce $\fmtvector{z}_i = \mathbb{E}[f](\fmtvector{q}_i)$.
In the following description $\fmtmatrix{K} \in \mathbb{R}^{N\times2}$ contains all the points $k_i$, $\fmtmatrix{V} \in \mathbb{R}^{N\times2}$ contains all the points $v_i$ and $\fmtmatrix{Q} \in \mathbb{R}^{N\times2}$ contains all the points $q_i$.

\paragraph{Models} 
\begin{itemize}
    \item \textbf{Intention}:
    \begin{equation}
        \begin{aligned}
        \fmtmatrix{E}_\fmtmatrix{K} &:= \fmtfunction{e}_\fmtmatrix{K}(\fmtmatrix{K}, \fmtmatrix{V}, \fmtmatrix{Q}) \\
        \fmtmatrix{E}_\fmtmatrix{V} &:= \fmtfunction{e}_\fmtmatrix{V}(\fmtmatrix{K}, \fmtmatrix{V}, \fmtmatrix{Q}) \\
        \fmtmatrix{E}_\fmtmatrix{Q} &:= \fmtfunction{e}_\fmtmatrix{Q}( \fmtmatrix{K}, \fmtmatrix{V}, \fmtmatrix{Q}) \\
        \fmtmatrix{w} &:= \fmtfunction{h}_\mathrm{int}( \fmtmatrix{E}_\fmtmatrix{K}, \fmtmatrix{E}_\fmtmatrix{V}, 
        \fmtmatrix{E}_\fmtmatrix{Q}) \\
        \fmtmatrix{\hat{P}}&:= \fmtfunction{e}_\fmtmatrix{P}(\fmtmatrix{w}) \\
        \end{aligned}
        \end{equation}

where $\fmtfunction{e}_\fmtmatrix{K}$, $\fmtfunction{e}_\fmtmatrix{V}$ and $\fmtfunction{e}_\fmtmatrix{Q}$ are two-layer MLPs with layer sizes [128, 128] and ReLU non-linearities.  $\fmtfunction{h}_\mathrm{int}$ is an Intention  module as described in the main paper and $\fmtfunction{e}_\fmtmatrix{P}$ is another MLP with sizes [128, 128, 2] and ReLU non-linearities.
To train the model we use Adam optimiser with a learning rate of $3e-4$.

\item \textbf{Attention}:
    \begin{equation}
        \begin{aligned}
        \fmtmatrix{E}_\fmtmatrix{K} &:= \fmtfunction{e}_\fmtmatrix{K}(\fmtmatrix{K}, \fmtmatrix{V}, \fmtmatrix{Q}) \\
        \fmtmatrix{E}_\fmtmatrix{V} &:= \fmtfunction{e}_\fmtmatrix{V}(\fmtmatrix{K}, \fmtmatrix{V}, \fmtmatrix{Q}) \\
        \fmtmatrix{E}_\fmtmatrix{Q} &:= \fmtfunction{e}_\fmtmatrix{Q}( \fmtmatrix{K}, \fmtmatrix{V}, \fmtmatrix{Q}) \\
        \fmtmatrix{w} &:= \fmtfunction{h}_\mathrm{MHA}( \fmtmatrix{E}_\fmtmatrix{K}, \fmtmatrix{E}_\fmtmatrix{V}, 
        \fmtmatrix{E}_\fmtmatrix{Q}) \\
        \fmtmatrix{\hat{P}}&:= \fmtfunction{e}_\fmtmatrix{P}(\fmtmatrix{w}) \\
        \end{aligned}
        \end{equation}

where $\fmtfunction{e}_\fmtmatrix{K}$, $\fmtfunction{e}_\fmtmatrix{V}$ and $\fmtfunction{e}_\fmtmatrix{Q}$ are two-layer MLPs with layer sizes [256, 256] and ReLU non-linearities.  $\fmtfunction{h}_\mathrm{int}$ is an Multihead Attention module with 5 heads and $\fmtfunction{e}_\fmtmatrix{P}$ is another MLP with sizes [256, 256, 2] and ReLU non-linearities.
To train the model we use Adam optimiser with a learning rate of $3e-4$.

\item \textbf{Neural Process}:
    \begin{equation}
        \begin{aligned}
        \fmtmatrix{E}_\fmtmatrix{H} &:= \fmtfunction{e}_\fmtmatrix{H}([\fmtmatrix{K}, \fmtmatrix{V}, \fmtmatrix{Q}]) \\
         \fmtmatrix{\hat{P}} &:= \fmtfunction{e}_\fmtmatrix{P}(\tfrac{1}{N}\sum_i\fmtmatrix{E}_{\fmtmatrix{H}_i})
        \end{aligned}
        \end{equation}

where $\fmtfunction{e}_\fmtmatrix{H}$ is a two-layer MLP with layer sizes [256, 256] and ReLU non-linearities. $\fmtfunction{e}_\fmtmatrix{P}$ is another MLP with sizes [256, 256, 2] and ReLU non-linearities.
To train the model we use Adam optimiser with a learning rate of $3e-4$.

\item \textbf{MLP}:
    \begin{equation}
        \begin{aligned}
         \fmtmatrix{\hat{P}} &:= \fmtfunction{e}_\fmtmatrix{P}([\fmtmatrix{K}, \fmtmatrix{V}, \fmtmatrix{Q}])
        \end{aligned}
        \end{equation}

where $\fmtfunction{e}_\fmtmatrix{P}$ is an MLP with layer sizes [256, 256, 256, 10] and ReLU non-linearity. 
To train the model we use Adam optimiser with a learning rate of $3e-4$.

\end{itemize}

\section{Anomaly detection}
\label{outlier}
For simplicity we use a ResNet34 pre-trained on ImageNet as a frozen representation extractor. Consequently we end up with images embedded as 512 dimensional vectors, 50,000 of which create we use to train our models and 10,000 left for testing. Splits are done over labels. Our networks use 4 heads with latent size of 256 each leading to hidden dimension of 1024. We also use input dropout of 20\% to counter overfitting. We train for 100 epochs with learning rate of 5e-5 and cosine warmup schedule over a period of 100 iterations.

Let us now define a single -former block with a single head.
\begin{itemize}
    \item Informer block

\begin{equation*}
\begin{aligned}
\fmtmatrix{E}_\fmtmatrix{X} &:= \max(0, e_\fmtmatrix{X}(\fmtmatrix{X}))\\
\fmtmatrix{E}_\fmtmatrix{K} &:= e_\fmtmatrix{K}(\fmtmatrix{E}_\fmtmatrix{K})\\
\fmtmatrix{E}_\fmtmatrix{V} &:= e_\fmtmatrix{V}(\fmtmatrix{E}_\fmtmatrix{V})\\
\fmtmatrix{E}_\fmtmatrix{Q} &:= e_\fmtmatrix{Q}(\fmtmatrix{E}_\fmtmatrix{X})\\
\alpha &:= \sigma_\mathrm{sigmoid}(\theta_\sigma)\\
\fmtmatrix{E}_\fmtmatrix{Z} &:= \fmtmatrix{E}_\fmtmatrix{Q}\fmtmatrix{E}_\fmtmatrix{K}'[(1-\alpha)\fmtmatrix{E}_\fmtmatrix{K}\fmtmatrix{E}_\fmtmatrix{K}' + \alpha \fmtmatrix{I}]^{-1} \fmtmatrix{E}_\fmtmatrix{V}
\end{aligned}
\end{equation*}
    
    \item $\sigma$Informer block

\begin{equation*}
\begin{aligned}
\fmtmatrix{E}_\fmtmatrix{X} &:= \max(0, e_\fmtmatrix{X}(\fmtmatrix{X}))\\
\fmtmatrix{E}_\fmtmatrix{K} &:= e_\fmtmatrix{K}(\fmtmatrix{E}_\fmtmatrix{K})\\
\fmtmatrix{E}_\fmtmatrix{V} &:= e_\fmtmatrix{V}(\fmtmatrix{E}_\fmtmatrix{V})\\
\fmtmatrix{E}_\fmtmatrix{Q} &:= e_\fmtmatrix{Q}(\fmtmatrix{E}_\fmtmatrix{X})\\
\alpha &:= \sigma_\mathrm{sigmoid}(\theta_\sigma)\\
\fmtmatrix{E}_\fmtmatrix{Z} &:= \sigma( \fmtmatrix{E}_\fmtmatrix{Q}\fmtmatrix{E}_\fmtmatrix{K}'[(1-\alpha)\fmtmatrix{E}_\fmtmatrix{K}\fmtmatrix{E}_\fmtmatrix{K}' + \alpha \fmtmatrix{I}]^{-1} )\fmtmatrix{E}_\fmtmatrix{V}
\end{aligned}
\end{equation*}
    
    \item Attention block

\begin{equation*}
\begin{aligned}
\fmtmatrix{E}_\fmtmatrix{X} &:= \max(0, e_\fmtmatrix{X}(\fmtmatrix{X}))\\
\fmtmatrix{E}_\fmtmatrix{K} &:= e_\fmtmatrix{K}(\fmtmatrix{E}_\fmtmatrix{K})\\
\fmtmatrix{E}_\fmtmatrix{V} &:= e_\fmtmatrix{V}(\fmtmatrix{E}_\fmtmatrix{V})\\
\fmtmatrix{E}_\fmtmatrix{Q} &:= e_\fmtmatrix{Q}(\fmtmatrix{E}_\fmtmatrix{X})\\
\fmtmatrix{E}_\fmtmatrix{Z} &:= \sigma( \fmtmatrix{E}_\fmtmatrix{Q}\fmtmatrix{E}_\fmtmatrix{K}')\fmtmatrix{E}_\fmtmatrix{V}
\end{aligned}
\end{equation*}
\item Linear Attention block

\begin{equation*}
\begin{aligned}
\fmtmatrix{E}_\fmtmatrix{X} &:= \max(0, e_\fmtmatrix{X}(\fmtmatrix{X}))\\
\fmtmatrix{E}_\fmtmatrix{K} &:= e_\fmtmatrix{K}(\fmtmatrix{E}_\fmtmatrix{K})\\
\fmtmatrix{E}_\fmtmatrix{V} &:= e_\fmtmatrix{V}(\fmtmatrix{E}_\fmtmatrix{V})\\
\fmtmatrix{E}_\fmtmatrix{Q} &:= e_\fmtmatrix{Q}(\fmtmatrix{E}_\fmtmatrix{X})\\
\fmtmatrix{E}_\fmtmatrix{Z} &:=  \fmtmatrix{E}_\fmtmatrix{Q}\fmtmatrix{E}_\fmtmatrix{K}' \fmtmatrix{E}_\fmtmatrix{V}
\end{aligned}
\end{equation*}
\item Neural Process Block
\begin{equation*}
\begin{aligned}
\fmtmatrix{E}_\fmtmatrix{Q} &:=e_\fmtmatrix{Q}(\fmtmatrix{Q})\\
\fmtmatrix{E}_\fmtmatrix{K} &:=e_\fmtmatrix{K}(\fmtmatrix{K})\\
\fmtmatrix{E}_\fmtmatrix{V} &:=e_\fmtmatrix{V}(\fmtmatrix{V})\\
\fmtmatrix{E}_\fmtmatrix{C} &:= e_\fmtmatrix{C}(\tfrac{1}{N} \sum_i [\fmtmatrix{E}_{\fmtmatrix{K}_i}; \fmtmatrix{E}_{\fmtmatrix{V}_i}])\\
\fmtmatrix{E}_{\fmtmatrix{D}_i} &:= [\fmtmatrix{E}_{\fmtmatrix{Q}_i}; \fmtmatrix{E}_\fmtmatrix{C}] \; \forall_i\\
\fmtmatrix{E}_\fmtmatrix{Z} &:= e_\fmtmatrix{Z}(\fmtmatrix{E}_\fmtmatrix{D})
\end{aligned}
\end{equation*}
    
\end{itemize}
Output of each head is then concatenated and followed by one more linear layer that embeds it in 1024 dimensional vector.
$e_\fmtmatrix{C}$ is an MLP with 1024 hidden units and 1024 output units.
$e_\fmtmatrix{X}$ is a linear layer shared between heads that maps into 1024 dimensional feature space. 
Remaining embeddings defined above are independent per head, and each one is also linear with 1024 outputs.

The above modules are used to define a layer of -former by applying
\begin{equation*}
\begin{aligned}
 \fmtmatrix{O}_1 & := 
 \mathrm{LayerNorm}( \fmtmatrix{E}_\fmtmatrix{Z} + \fmtmatrix{X} ) \\
 \fmtmatrix{O}_2 & :=
 \mathrm{LayerNorm}(e_\fmtmatrix{O}(\fmtmatrix{O}_1) + \fmtmatrix{O}_1) 
\end{aligned}
\end{equation*}
where $e_\fmtmatrix{O}$ is a an MLP with a hidden size of 2048 and output of 1024.

We stack either 1 or 4 of these layers (each with its own learnable parameters) to form Transformer, Informer, Linear Transformer, NP-former and $\sigma$Informer.

We use softmax to produce probability estimates at the output and a cross entropy loss to train the model.

% Both \infor{} and $\sigma$\infor{} use a trainable $\alpha$ parameter and a parametrisation of smoothing through a convex combination of identity and empirical covariance
% $
% \sigma_\mathrm{sigmoid}(\alpha) \fmtvector{K}\fmtvector{K}' 
% +
% (1-\sigma_\mathrm{sigmoid}(\alpha))
% \fmtvector{I}.
% $

\section{Normalisation}
One of the elements introduced in the Transformer paper~\citet{vaswani2017attention} has been use of scaled attention
$$
\tfrac{1}{\sqrt{d}} \fmtmatrix{Q}\fmtmatrix{K}',
$$
in order to preserve variance of 1 of the attention mask at initialization. For the reminder of this section we assume both queries and keys are random vectors coming from a standard normal distribution. We also focus on the part of the computation that ignores values, as in the transformer paper.
Given that we have proven Intention behaves like attention in the limit it is natural to expect it having similar properties. 

First, let us investigate a simple case of $N=d=1$ meaning we just have a single key and it is just a number. Consequently intention computation becomes just
$$
\fmtscalar{q}\fmtscalar{k}(\fmtscalar{k}\fmtscalar{k})^{-1} = \fmtscalar{q} \cdot \tfrac{1}{\fmtscalar{k}}.
$$
Both $\fmtscalar{q}$ and $\fmtscalar{z} := \tfrac{1}{\fmtscalar{k}}$ are independent random variables, with $\fmtscalar{q} \sim \mathcal{N}(0,\fmtscalar{1})$. On the other hand $\fmtscalar{z}$ follows reciprocal normal distribution. Despite relative simplicity this distribution does not have a well defined expectation nor variance~\cite{johnson1994continuous}. 

While this result might seem very bad, we note that in practise we never work with such small dimensions in deep learning, and in practise these are unlikely to be much of an issue and thus look at a more generic scaling behaviour with $N=1$ and $d \rightarrow \infty$.
$$
\fmtvector{q}\fmtvector{k}'(\fmtvector{k}\fmtvector{k}')^{-1} = \fmtvector{q} \cdot \tfrac{\fmtvector{k}'}{\|\fmtvector{k}\|^2}.
$$
Since norm of a random normal vector is approximately $\sqrt{d}$ we see that it will converge to 0 as the dimension grows, potentially collapsing representation in intention module to a single point.
However, this also suggests that in order to recover a good scaling behaviour we can just counter this effect by multiplying the whole computation by $\sqrt{d}$ (as opposed to attentions \emph{division} by the same factor). 

\begin{figure}[ht]
\begin{center}
\centerline{\includegraphics[width=\columnwidth]{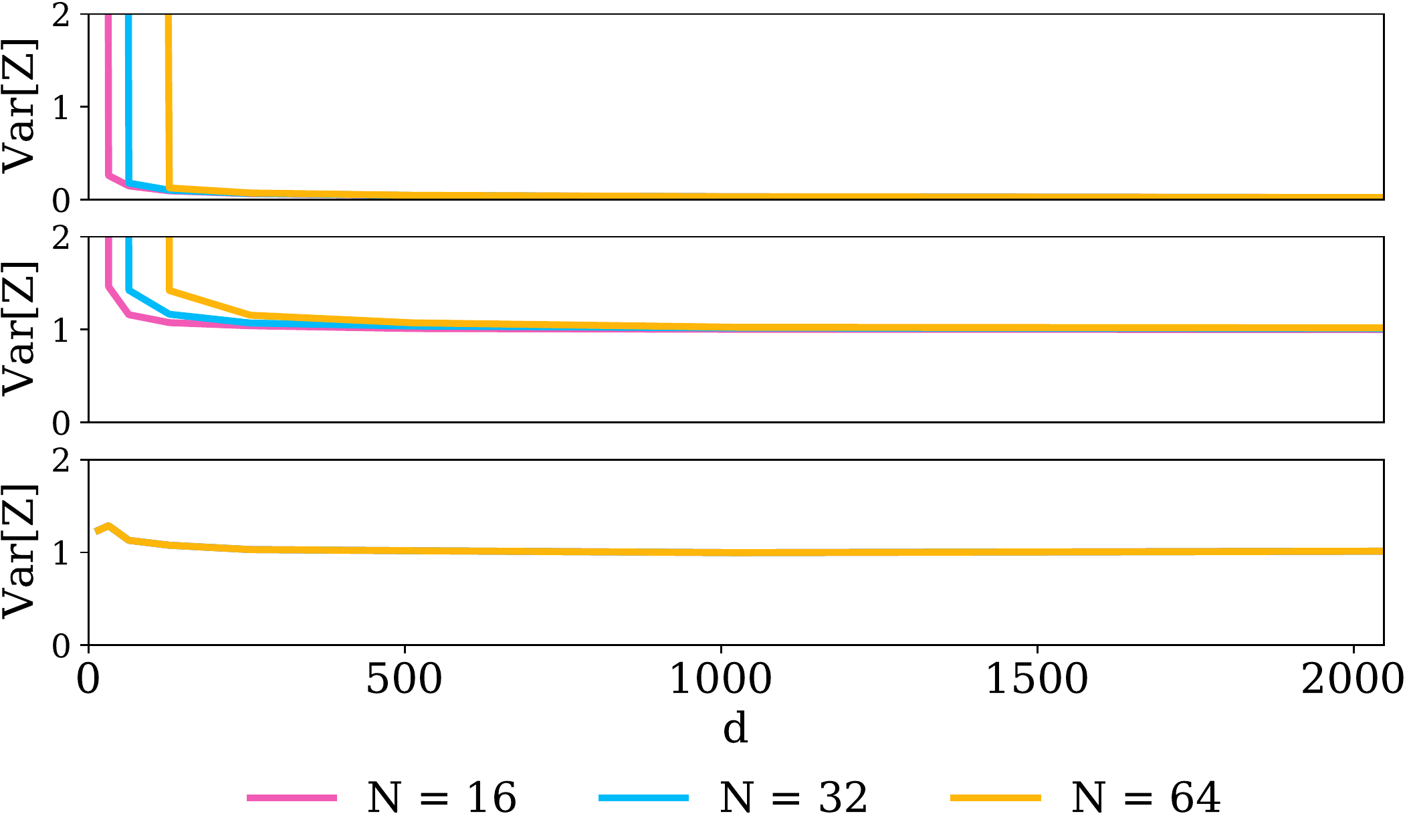}}
\caption{Comparison of variances of the Intention map $\fmtmatrix{Z}$ when both keys and queries come from $d$ dimensional random normal distribution with $N$ points. From top: using unscaled Intention, scaled Intention and scaled Intention with a regulariser.}
\label{fig:variances}
\end{center}
\vskip -0.2in
\end{figure}

We compare variance of $\fmtmatrix{Z}$ empirically by taking $N=16,32,64$ points in growing number of dimensions and applying:
\begin{itemize}
    \item Unscaled Intention $$\fmtmatrix{Z} := \fmtmatrix{Q}\fmtmatrix{K}'[\fmtmatrix{K}\fmtmatrix{K}']^{-1}$$
    \item Scaled Intention $$\fmtmatrix{Z} := \sqrt{d} \fmtmatrix{Q}\fmtmatrix{K}'[\fmtmatrix{K}\fmtmatrix{K}']^{-1}$$
    \item Scaled Intention with a regulariser
    $$\fmtmatrix{Z} := 0.5\sqrt{d} \fmtmatrix{Q}\fmtmatrix{K}'[\sigma_\mathrm{sig}(\alpha)\fmtmatrix{K}\fmtmatrix{K}' + (1-\sigma_\mathrm{sig}(\alpha))\fmtmatrix{I}]^{-1},$$
    where $\alpha$ is initialised to 0 (trainable parameter).
\end{itemize}
We can see in Figure~\ref{fig:variances} that as expected unscaled intention has an exploding variance for very small $d$ and a vanishing representation problem as $d$ goes to infinity. Scaled Intention smoothly converges in variance to 1, and having a regulariser stabilises behaviour for small $d$ too.
We note, that this is an optional element, and mostly matter for Informers, rather than using an Intention module as a neural network head. In our experiments we successfully trained modules up to 4-6 layers even with unscaled Intention. One can attribute it to existence of multiple other mechanisms in Deep Learning that counter issues emerging here, such as LayerNorm layers etc.

% \end{toappendix}
% \bibliography{example_paper}
% \bibliographystyle{icml2022}

% \bibliography{example_paper}
% \bibliographystyle{icml2022}

\end{document}